\documentclass[sigconf]{acmart}

\AtBeginDocument{%
	}

\copyrightyear{2024}
\acmYear{2024}
\setcopyright{rightsretained}
\acmConference[FAccT '24]{The 2024 ACM Conference on Fairness, Accountability, and Transparency}{June 3--6, 2024}{Rio de Janeiro, Brazil}
\acmBooktitle{The 2024 ACM Conference on Fairness, Accountability, and Transparency (FAccT '24), June 3--6, 2024, Rio de Janeiro, Brazil}\acmDOI{10.1145/3630106.3658993}
\acmISBN{979-8-4007-0450-5/24/06}

\acmSubmissionID{13794.108}

\begin{document}
	
	\title{\emph{A Robot Walks into a Bar}: Can Language Models Serve as Creativity Support Tools for Comedy?\\An Evaluation of LLMs' Humour Alignment with Comedians}
	
	\author{Piotr W. Mirowski}
	\authornote{Both authors contributed equally to this research.}
	\email{piotrmirowski@deepmind.com}
	\orcid{0000-0002-8685-0932}
	\affiliation{%
		\institution{Google DeepMind}
		\streetaddress{5 New Street Square}
		\city{London}
		\country{UK}
	}
	
	\author{Juliette Love}
	\authornotemark[1]
	\email{juliettelove@deepmind.com}
	\affiliation{%
		\institution{Google DeepMind}
		\streetaddress{5 New Street Square}
		\city{London}
		\country{UK}
	}
	
	\author{Kory Mathewson}
	\orcid{0000-0002-5688-6221}
	\email{korymath@deepmind.com}
	\affiliation{%
		\institution{Google DeepMind}
		\streetaddress{5 New Street Square}
		\city{Montréal, QC}
		\country{Canada}
	}

	\author{Shakir Mohamed}
	\orcid{0000-0002-1184-5776}
	\email{shakir@deepmind.com}
	\affiliation{%
		\institution{Google DeepMind}
		\streetaddress{5 New Street Square}
		\city{London}
		\country{UK}
	}
	
	\renewcommand{\shortauthors}{Mirowski et al.}
	
	\begin{abstract}
		We interviewed twenty professional comedians who perform live shows in front of audiences and who use artificial intelligence in their artistic process 
as part of 3-hour workshops on ``AI x Comedy'' conducted at the Edinburgh Festival Fringe in August 2023 and online.
The workshop consisted of a comedy writing session with large language models (LLMs), a human-computer interaction questionnaire to assess the Creativity Support Index of AI as a writing tool, and a focus group
interrogating the comedians' motivations for and processes of using AI, as well as their ethical concerns about bias, censorship and copyright.
Participants noted that existing moderation strategies used in safety filtering and instruction-tuned LLMs reinforced hegemonic viewpoints by erasing minority groups and their perspectives, and qualified this as a form of censorship. At the same time, most participants felt the LLMs did not succeed as a creativity support tool, by producing bland and biased comedy tropes, akin to ``cruise ship comedy material from the 1950s, but a bit less racist''. Our work extends scholarship about the subtle difference between, one the one hand, harmful speech, and on the other hand, ``offensive'' language as a practice of resistance, satire and ``punching up''. We also interrogate the global value alignment behind such language models, and discuss the importance of community-based value alignment and data ownership to build AI tools that better suit artists' needs. {\bf Warning: this study may contain offensive language and discusses self-harm.}


	\end{abstract}


	
	\received{22 January 2024}
	
	\maketitle
	
	\section{Introduction}
	\subsection{Motivation: investigate the potential and implications of LLMs for comedy writing}


Recent work on the intersection of AI and comedy has demonstrated \citep{toplyn2023witscript3,chen2023prompt,jentzsch2023chatgpt,winters2021computers,mirowski2019human,mathewson2018improbotics} an appetite for comedians to (try to) write humorous material using AI tools like Large Language Models (LLMs). We conducted an empirical study to better understand the current state of LLMs as comedy-writing support tools, their use-cases and limitations, and artists' opinions on ethical questions regarding their use in a comedy-writing context. The complexity of comedy can help expose some limitations of LLMs. To participate, we recruited 20 professional comedians who use AI in their artistic processes and who perform live shows in front of audiences (10 in person at the Edinburgh Festival Fringe, 10 online) for a 3-hour-long workshop on ``AI x Comedy''. As detailed in Section \ref{sec:methods}, the workshop consisted of a comedy writing session with generally-available instruction-tuned LLMs (ChatGPT \citep{openai2023gpt4,ouyang2022training} and Bard \citep{chowdhery2023palm,geminiteam2023gemini}), a human-computer interaction questionnaire, and focus group discussions on the use of LLMs in comedy writing and ethical concerns.

\subsubsection{Using LLMs for humour, a task with human-level difficulty}
\label{sec:fascination}

Trying to combine humour and machine intelligence is a long-standing subject of scientific enquiry \citep{sharples2022story,veale2021your,raskin1979semantic}, and is perceived as a fundamental challenge. According to computational humour researchers like \citet{winters2021computers}, ``humans are the only known species that use humor for making others laugh''
\citep{caron2002ethology,gervais2005evolution}.
\citet{winters2021computers} argues that one of the modern formal humor theories points to \emph{incongruity} \citep{hutcheson1750reflections} (whereby the setup points in one direction and the punch line in another) as a basic element \citep{gervais2005evolution,ritchie1999developing,raskin1979semantic}\footnote{Alternative humour theories include the Aristotelian \emph{Relief Theory} \citep{aristotle350BCpoetics} of tension and release whereby we let out our psychic energy connected with repressed topics, the \emph{Superiority Theory} \citep{hobbes1651leviathan} whereby we laugh at others’ misfortunes to feel better about ourselves, and the \emph{Benign Violation Theory} \citep{mcgraw2010benign}.}. As we discuss in Section \ref{sec:human}, producing and resolving incongruity is a task with human-level difficulty. We situate LLMs for comedy within broader computational humour research and AI-assisted comedy performance \citep{mathewson2017improvised,rosa2020theaitre,mirowski2023co} in Appendix \ref{sec:computational-comedy}.



\subsubsection{The utility of instruction-tuned LLMs as Creativity Support Tools}
Similar to previous empirical studies on the use of LLMs for creative writing \citep{qadri2023ai,mirowski2023co,chakrabarty2023creativity,chakrabarty2023art,huang2023inspo,chakrabarty2022help,yuan2022wordcraft,gero2022sparks,calderwood2022spinning,ippolito2022creative}, we asked the artists about their motivations and processes for using LLMs. We asked about the potential and limitations of language models as \emph{Creativity Support Tools} \citep{chakrabarty2023creativity}, and quantified the \emph{Creativity Support Index} of LLMs for comedy writing \citep{cherry2014quantifying}. We report our results in Section \ref{sec:results-quantitative}.


\subsubsection{Socio-Technical Systems concerns with LLMs for creative writing}
\label{sec:socio-technical}

Inspired by \citet{dev2023building}, we leveraged community engagement with large generative models, and interrogated the diverse, intersectional identities of the comedians using AI in a creative context. In addition to their reasons for using (or not using) LLMs as comedy-writing tools, we asked participants about the ethical considerations of using AI. The study was conducted at the time of the Writers' Guild of America (WGA) 2023 strike \citep{wga2023summary}. Participants raised and addressed questions on the scrutiny of AI and on concerns around AI's impacts, both on intellectual property and artistic copyright, and on artists' livelihoods. We report their opinions in Section \ref{sec:results-qualitative} and discuss these concerns in Section \ref{sec:copyright-jobs}.

\subsection{Investigating hypotheses about using LLMs to write comedy}
\label{sec:expected}

Based on previous work and on the authors' personal experience of AI as creativity support tools, we hypothesized---prior to conducting our study---that participants would express negative opinions of LLMs for co-creativity on four issues: expressing stereotyped (Sect. \ref{sec:bias}) or bland (Sect. \ref{sec:homogeneisation}) language, censorship (Sect. \ref{sec:offensive}) and missing context (Sect. \ref{sec:missing-context}). We review literature on these four hypotheses below, as they become the basis of our mixed methods study.

\subsubsection{Biases in large language models}
\label{sec:bias}

Gender and racial biases embedded within machine learning models have been extensively documented \citep{buolamwini2018gender,hamidi2018gender,benjamin2019race,west2019discriminating,abid2021persistent,bender2021dangers,blodgett2021stereotyping,bommasani2021opportunities,sheng2021societal,dev2022measures}. These biases include sexism and racism \citep{bender2021dangers,sheng2021societal}, homophobia and transphobia \citep{queerinai2023queer,dias2021fighting}, Islamophobia \citep{abid2021persistent}, the perpetuation of Western colonial mindsets \citep{mohamed2020decolonial}, Anglocentrism \citep{zhou2021frequency}, and in-group vs. out-group social identity biases \citep{hu2023generative}.
In their extensive reviews, \citet{bommasani2021opportunities,rauh2022characteristics} identified two broad kinds of harms resulting from such biases: \emph{intrinsic harms} such as representational bias (due to misrepresentation, overrepresentation, and underrepresentation of specific social groups), and \emph{extrinsic harms}, the downstream consequences of biased models, including representational and performance disparities. As we show in Section \ref{sec:results-qualitative}, the study participants noticed a few examples of representational harm and many examples of underrepresentation harm (also called \emph{allocational harm} in \citep{rauh2022characteristics}), such as erasure when LLMs refused to generate content for certain demographic groups.

\subsubsection{Potential censorship of speech labeled as ``\emph{offensive}''}
\label{sec:offensive}

Comedians often pepper their language with profanities and their material with provoking themes. As we discuss in Section \ref{sec:hhh} (and confirmed by the study participants in Section \ref{sec:results-qualitative}), \emph{offensive} language that would be perfectly acceptable at a comedy club may get ``censored'' by instruction-tuned LLMs that ``refuse'' to answer ``offensive'' prompts.

This problem has been observed in automated moderation of online content, such as hate speech detectors that suppress social media posts by queer communities and drag queens \citep{dias2021fighting,diaz2022accounting}, or posts using African-American Vernacular English \citep{amironesei2023relationality,xu2021detoxifying}. \citet{amironesei2023relationality} called it \emph{censorship}. \citet{rauh2022characteristics} studied algorithmic moderation of social media posts by the Perspective API, noting that ``authors of the comment may be harmed if their content is incorrectly flagged as toxic'' by the moderation tool. Similar erasure due to the cultural hegemony embedded in image generators has been studied in \citep{qadri2023ai}. We relate the participants' experience, similarly frustrated that the LLM tools ``considered'' their own identity and comedy material as problematic and necessary to censor.

\citet{diaz2022accounting} defined \emph{offensive language} as non-normative ``language that uses terminology that is noted as offensive but which is not perceived as offensive in particular contexts of use'', and studied its use by minorities as a form of resistance, for ``socially productive uses of decoratively offensive language'', aiming to reclaim ``offensive'' language and resist oppression. Just like the minority groups described in \citep{diaz2022accounting}, many comedians (who may be members of minority groups themselves) often use offensive jokes to \emph{punch up}, and satire (``to challenge existing social structures'') to build empathy, rather than to \emph{punch down} (``silence others'').

\subsubsection{Missing context}
\label{sec:missing-context}

Context is key to disambiguate offensive language from hate speech. LLMs, like social media posts, cause ``context collapse'' \citep{marwick2011tweet} by providing a limited amount of information to understand their meaning, particularly when using mock impoliteness. Specifically, ``in-group usage of reclaimed slurs can be considered acceptable, depending on who uses them'' \citep{rauh2022characteristics,croom2011slurs}. Moreover, the context of comedy extends beyond the language to other factors including the audience and the venue.

\subsubsection{Homogeneisation}
\label{sec:homogeneisation}

\citet{bommasani2021opportunities} warned that ``the application of foundation models across domains has the potential to act as an epistemically and culturally homogenising force, spreading one perspective, often a socially dominant one, across multiple domains of application''. In the arts, this means that AI-generated artifacts may lead to a homogeneisation of aesthetic styles \citep{weidinger2023sociotechnical,epstein2023art}, further reinforced by curation algorithms \citep{epstein2023art,manovich2018ai}. For creative writing, empirical studies showed that instruction-tuned LLMs reduce the diversity of content in co-writing tasks \citep{padmakumar2023does}, and that LLM-generated stories did not pass the \emph{Torrance Test of Creative Writing} according to metrics of ``fluency, flexibility, originality and elaboration'' \citep{chakrabarty2023art}. Similarly to \citet{qadri2023ai}, we ran focus groups with artists to interrogate cultural (Western) biases (see Sections \ref{sec:focus-group-questions} and \ref{sec:focus-group-analysis}).

\subsubsection{Investigating hypotheses via a mixed-methods study}
In our empirical study, we ask participants questions on all four problems identified in Section \ref{sec:expected}, namely about bias, censorship, context and homogeneity. In Section \ref{sec:discussion}, we build upon scholarship on cultural value alignment of language models, the moderation of offensive and harmful speech, and the use of offensive speech and satire as a form of resistance, to revisit the global cultural value alignment of LLMs and propose community-based alignment to build LLMs that better suit comedians' creative needs\footnote{We deliberately do not generalize our findings beyond comedy, as some professionals and the computational creativity community have historically embraced LLM tools in a way that fits their creative practice, whether building on the glitch aesthetic \citep{manovich2018ai,parrish2017poetic} or designing interactive experiences \citep{park2023generative}.}.
	
	\section{Methods}
	\label{sec:methods}

Our study was designed to address a challenging problem, with on one hand, limitations of LLMs (stereotypes, inability to distinguish comedic offensiveness from harmful speech, cultural erasure and homogeneisation of content), and on the other hand, the use of LLMs for a creative writing task. For this reason, we asked a group of experts---professional comedians and performers---who are used both to thinking about thorny questions of identity, offensiveness and censorship in their work, and to employing language in a highly creative way. We chose artists who already use AI in their work and expected them to be somewhat knowledgeable and open to using AI: this likely biased our results\footnote{Our biased selection criteria of participants might, and likely do, lead to biased opinions as compared to the much more broad population of comedians and performers, which might be reflected in more favourable judgment of the Creativity Support Index of LLM writing tools. Future research might explore the diversity of opinions in creative communities across a greater range of familiarity with AI tools and openness to using them in their own creative practices. Exploring those opinions would significantly increase the scope of the paper and would make a compelling follow-up study.}.

We ran workshops with 20 comedians who use AI creatively. The first workshop with 10 participants was run in person at Edinburgh Festival Fringe 2023; the following 3 workshops with 3, 4 and 3 participants were run online.
We reached out to comedians performing in Edinburgh during Fringe, or in our network, and attempted to recruit as diverse (along linguistic, cultural, gender, sexual, national and racial dimensions) a pool of comedians as possible given the constraints of the study\footnote{A demographic analysis of opinions might be a possible avenue for future investigations, but it would require a different study design and participant recruiting process.}. Participants had contrasting views on AI for comedy writing, from "AI is very bad at this, and I don't want to live in a world where it gets better" (p15) to "I liked the details that I got. I think those details sparked my imagination, and I think I could use them to write something" (p20).
Participants were asked to register on the Prolific platform\footnote{\url{https://prolific.com}} and invited to join a specific study thanks to an allowlist. The study was approved by the research ethics committee of our institution. The information sheet and consent forms were shared with the participants, their active consent was obtained at the beginning of the workshop and they had the right to withdraw without prejudice at any time. The Prolific platform handled the payment of their participation fee, set to £300 for 3 hours.

We started each 3-hour session by describing the agenda and goals of the workshop, sharing the information sheet and consent forms with the participants, and asking them to start filling out a short anonymous survey about their background in comedy, previous exposure to AI and usage of AI in performance (full questionnaire in Appendix \ref{app:past}).

\subsection{Writing exercise}
\label{sec:writing}

We then proceeded with a comedy-writing exercise, in which participants spent around 45 minutes on their own, using an LLM. We encouraged participants to try to use the LLM in a way that would generate useful material ``that they would be comfortable presenting in a comedy context'', but emphasized that we did not require a fully-finished product by the end of the writing exercise. We invited them to use the language(s) they felt the most comfortable with\footnote{Languages included German, Dutch, English, French, Hindi, Swedish and Tamil.}.
We also suggested they could use the tool to 1) generate, rate/detect or explain jokes, 2) co-write jokes via iterative prompting, step-by-step or using examples, and 3) analyse, re-write or complete some of their previous material. In the first workshop (in person), we provided participants with access to ChatGPT-3.5 \citep{ouyang2022training} served via a plain text interface similar to ChatGPT. In the following 3 workshops, we invited participants to use their own preferred model via their personal account: participants used ChatGPT-3.5, ChatGPT-4 \citep{openai2023gpt4} and Google Bard powered by  Gemini Pro \citep{geminiteam2023gemini} (December 2023 version).
Note that the choice of such instruction-tuned models was motivated by their popularity and ease of access by comedians, and more complex prompting strategies, such as used in Dramatron \citep{mirowski2023co}, could have produced higher-quality outputs.

\subsection{Creativity Support Tools evaluation}
\label{sec:csi}

Following the writing exercise, we asked participants to fill out three surveys. The first survey was about their experience with the AI system for writing comedy material and contained nine questions from previous studies \citep{ippolito2022creative,mirowski2023co,yuan2022wordcraft} that assessed LLMs for creative writing on the 5-level Likert scale (see Appendix \ref{app:experience}). The second survey was used to calculate the Creativity Support Index (CSI) \citep{cherry2014quantifying} of the writing tool, which itself was adapted from the NASA Task Load Index \citep{hart1988development}. CSI is estimated in a psychometric survey that measures six dimensions of creativity support: \emph{Exploration}, \emph{Expressiveness}, \emph{Immersion}, \emph{Enjoyment}, \emph{Results Worth Effort}, and \emph{Collaboration} (see specific questions in Appendix \ref{app:csi}), and is a number between 0 and 100, where 90 is considered excellent and 50 mediocre. The third survey contained free-form questions on one thing that the ``AI system'' (the LLM writing tool) did well, one improvement, and open-ended comments on the writing session and on the survey.

\subsection{Focus Group Questions}
\label{sec:focus-group-questions}

The last part of the workshop consisted in a one-hour focus group, where we asked participants first to discuss the writing task (for about 30 minutes) and second to discuss the general usage of AI for writing comedy material.

In order to guide the discussion, we prepared two sets of questions\footnote{Question-led focus groups are useful to start discussions, but we acknowledge the limitation that questions can bias the participants' responses.} (see Appendix \ref{app:free} for the full list of questions). The first set of questions pertained to the usefulness of the outputs generated by the LLM tool for personal writing, differences between using an LLM or searching for inspiration using Wikipedia or a search engine, the types of comedy that can be produced by an LLM, and concerns about the ownership of LLM-generated outputs.

The second set of questions addressed the comedy writing process of the participants, as well as the topics introduced in Section \ref{sec:expected}, namely various biases and stereotypes of LLMs, problems with moderation strategies employed by LLMs, the importance of context and delivery or whether some forms of cultural appropriation or homogeneisation could happen. We invited discussions about the use of other comedians' work, and also challenged the participants with question on whether the AI has a ``voice'' and if humour can be quantified.

\subsection{Focus Group Analysis}
\label{sec:focus-group-analysis}

In our workshops, we had followed focus group methodology described in \citep{noble2016grounded,onwuegbuzie2009qualitative} (engaging a group of participants in an informal one hour discussion focused around a particular topic, activity, or stimulus material, with a team of two moderators). Transcripts of focus groups were recorded as audio recordings, then automatically transcribed using speech recognition tools in Google Meet, and manually verified as well as compared against notes taken by the moderators. After transcription, audio and video recordings were destroyed. Like in the surveys, participants were anonymised: authors independently reviewed the transcripts to remove any personally identifiable information from the transcripts.
We then performed \emph{constant comparison analysis} to analyze the transcripts of the focus groups \citep{onwuegbuzie2009qualitative}. We first identified initial codes using sentence-by-sentence open coding. We then grouped those codes into themes, and found themes that were coherent across focus groups. Data from four focus groups allowed us to achieve \emph{data saturation} \citep{braun2006using,maguire2017doing}.


Results section \ref{sec:results-quantitative} summarises the quantitative results\footnote{
Full outputs of the writing sessions, all individual survey results and raw transcripts from the focus groups will be shared in anonymised form as supplementary material, once our work is published.} derived from the Creativity Support Tool evaluation (Sect. \ref{sec:csi}), while results section \ref{sec:results-qualitative} details the observations made by the participants during focus groups (Sect. \ref{sec:focus-group-analysis}).
Please note that this paper is an exploration of external perspectives rather than an endorsement of any one of them; in particular, this paper does not seek to undertake any legal evaluation.
	
	\section{Quantitative Results and Creativity Support Index}
	\label{sec:results-quantitative}

\begin{figure}[h]
  \centering
  \includegraphics[width=0.85\linewidth]{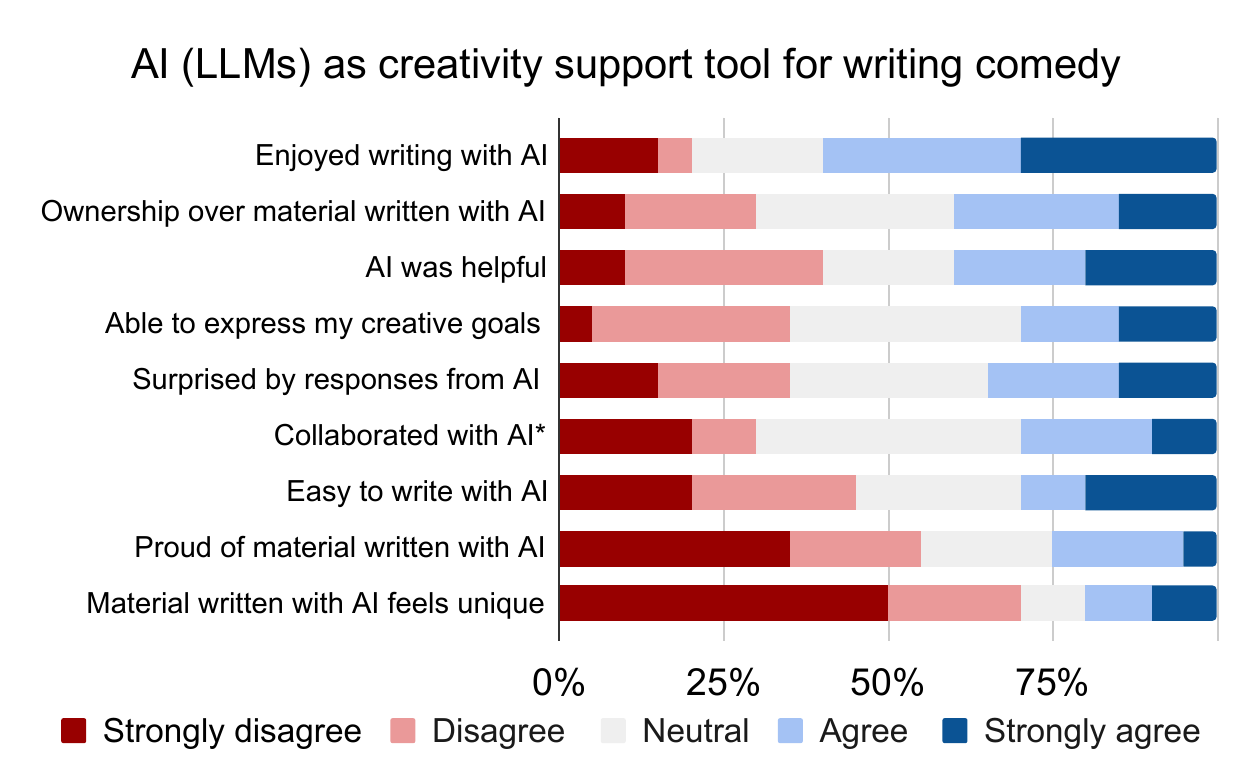}
  \includegraphics[width=0.90\linewidth]{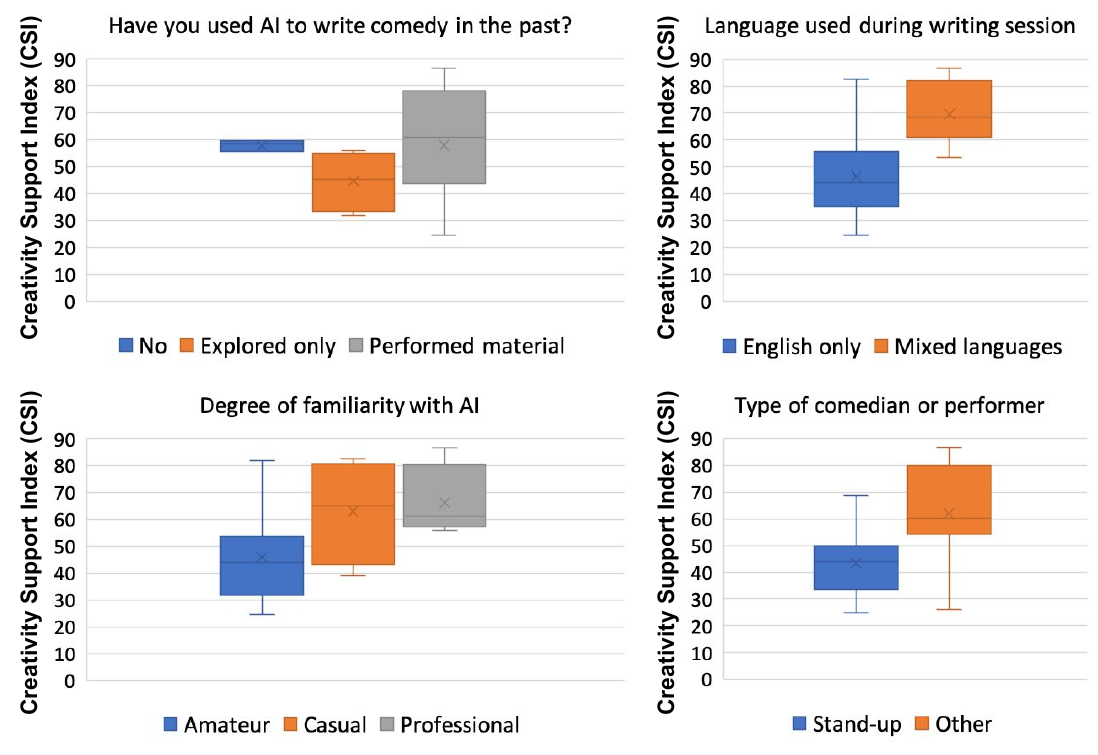}
  \caption{Left: Evaluation of the instruction-tuned LLMs as creativity support tool for writing comedy using a Likert scale; each row corresponds to a question in a survey (questions are listed in Appendix \ref{app:experience}). Only 10 participants responded to question ``Collaborated with AI''. Right box plots (box for quartiles and whiskers for min and max) show the break-down of the Creativity Support Index \citep{cherry2014quantifying}, respectively by a) previous usage of AI for writing comedy material, b) language used during the writing session with LLMs, c) the degree of familiarity with AI and d) the type of comedian or performer (stand-up vs. improv, theatre, film, dance etc.).}
  \label{fig:results}
  \Description{Plots showing the quantitative results. Left: Evaluation of the instruction-tuned LLMs as creativity support tool for writing comedy using a Likert scale; each row corresponds to a question in a survey (the full list of questions is listed in Appendix). Right plots (box for quartiles and whiskers for min and max) show the break-down of the Creativity Support Index \citep{cherry2014quantifying}, respectively by a) previous usage of AI for writing comedy material, b) language used during the writing session with LLMs, c) the degree of familiarity with AI and d) the type of comedian or performer (stand-up vs. improv, theatre, film, dance etc.).}
\end{figure}

Figure \ref{fig:results} summarises the quantitative results collected at the end of the writing session with instruction-tuned LLMs. Based on the participants' responses to the survey questionnaire (see Appendix \ref{app:experience} for questions), we notice that while comedians mostly enjoyed writing with AI, questions about the ownership, helpfulness, expressivity, surprise, collaboration and ease of writing with AI had mixed answers. The majority of participants did not feel pride in the material written with AI, nor did they feel it was unique, hinting at the derivative nature of AI-generated text.

We computed the Creativity Support Index (see Appendix \ref{app:csi} for computation), with an average score that is mediocre ($\mu=54.6$, $\sigma=18.1$). We broke down CSI based on participants' background in comedy, previous exposure to AI and usage of AI in performance (see questions in Appendix \ref{app:past}), as well as based on the language they used during the writing session. As Figure \ref{fig:results} shows, participants who scored the LLMs writing tools highest were those who had professional exposure to AI, or who had performed AI-written material in the past (as opposed to only exploring the tool), or who qualified themselves as improvisers, theatre and film actors and directors, or dancers, rather than as full-time stand-up comedians. Interestingly, participants who used LLMs to generate non-English or multi-lingual text scored the tool higher on CSI. Lastly, we did not observe a significant change in CSI among the models between August 2023 ($\mu=50$, $\sigma=15.8$) and December 2023 ($\mu=60$, $\sigma=19.6$).

	\section{Qualitative Results from Focus Group Discussions}
	\label{sec:results-qualitative}

In this section, we summarize the major themes that emerged from the focus group discussions.
Each is presented alongside supporting quotes from various participants (anonymised as p1, p2, etc.).

\subsection{Use-cases of LLMs in comedy writing and quality of generated outputs}
Participants described a diverse array of use cases for LLMs in their writing practice, including as a conversational brainstorming partner (p19), critic (p6), choreographic assistant (p3), translator (p1, p11, p12, p19, p20), and historical guru (p13). They generally reflected positively on the potential of LLMs to assist with some tasks within the comedy writing process. However, many participants commented on the overall poor quality of generated outputs, and the amount of human effort required to arrive at a satisfying result.

\subsubsection{LLMs can be an effective first step for quickly generating content and structure.}
Participants described the utility of LLMs for generating content much faster than human writers. They described success with using LLMs to generate first drafts, which then required significant edits from human writers: ``AI allows you to kind of get that s*** first draft immediately'' (p6). Participant p14 called their initial output ``a vomit draft that I know that I'm gonna have to iterate on and improve.'' Many participants also described using LLMs to generate a structure for a sketch or other performance, of which they could then fill in the details—the LLM ``spat out a scene which provided a lot of structure'' (p17).

\subsubsection{Generated outputs are generally of poor comedic quality.}
Many participants noted that they only used LLMs for setup and structure generation due to their inability to generate humorous outputs from the models:
``the most bland, boring thing---I stopped reading it. It was so bad'' (p6),
``just consistently bad [...] didn't really improve on the jokes'' (p10).
Some participants described particular aspects critical in comedy, and how LLMs seemed incapable of them: ``AI generated material has a lack of agency.
[...] lacking that little bit of urgency that shows it can be emotional'' (p11).

\subsubsection{Participants had difficulty steering LLMs away from bland and generic outputs.}
Six participants described LLM-generated outputs as ``bland'' or ``generic,'' making them poor producers of comedic or artistic material: ``the words seem very generic. They lack that incisiveness that I often find with human written language'' (p11);
``if you zoom out on the story that it told, it wasn't really a good story or a creative story'' (p20). Participants also commented on various prompting approaches and their general lack of success at prompting the LLM to generate more specific or interesting responses: ``no matter how much I prompt [...] it's a very straight-laced, sort of linear approach to comedy'' (p11).

\subsubsection{The human writer still produces the humorous elements in co-written text.}
The importance of human writers in providing the comedic aspects of material written with LLMs was a common theme. Many participants commented that while LLMs could provide effective setup or structure, they often could not provide the humor: ``usually it can serve in a setup capacity. I more often than not provide the punchline'' (p17). When participants had success using LLMs in the writing process, they still attributed the best parts of produced output to the human in the loop: ``the only thing again that is funny in what I gave you is the joke I put into the prompting'' (p15).

\subsubsection{Lack of concern over ownership over generated content}
In response to questions about feelings of ownership over content that was co-written with LLMs, most participants felt little concern. For some, this was due to the poor quality of generated outputs: ``most of the jokes I was writing [are] the level of, I will go on stage and experiment with it, but they're not at the level of, I'd be worried if anyone took one of these jokes'' (p14). For others, it was due to the amount of human effort required in improving  generated outputs: ``I don't feel a lot of ownership, because there's no finished product. If I could polish it, then it would feel more like it is mine'' (p19).

\subsection{Limitations introduced by moderation and safety filtering}
Participants commented on the moderation and safety filtering applied to widely-available language models. They remarked that this moderation limited the creative agency of human writers using LLMs, by serving as an initial editor of the text and removing writers’ ability to self-moderate. They also expressed frustration at being unable to use LLMs to write about many themes common in comedy writing, including sexually-suggestive material (p3, p10, p13, p16), dark humor (p8), and offensive jokes (p10, p15, p20).

\subsubsection{Moderation and safety filtering limits writers’ creative agency.}
Participants explained that self-moderation is a critical part of the writing process, and expressed frustration that moderation tools on LLMs interfered with that process: ``the creative process is about going through stages of ‘this material isn't good enough, it's not right, or it's offensive, it's marginalizing people, I need to make it more acceptable.’ And I think AI models are a beginning to do that before you have a chance to explore'' (p10). ``It probably would be more interesting for a writer if there would be less moderation, because you can do the moderation in your own prompts. A writer is going to moderate themselves. If you're writing with an AI, if you don't like the bad stuff that it writes, you won't use it'' (p19). Participants described that this external source of moderation limited the creative control of the human writer: ``it's interesting if there's less moderation, because... the end result is moderated in the way that the author wants it to be'' (p19). Some explained that if filters were necessary, that the user should still have some degree of control over them: ``I feel like the opportunity to set the filters should still be at the performers’ end'' (p12).

\subsubsection{Moderation limits writers’ ability to use AI with their preferred subject matter.}
In addition to affecting the creative writing process, participants commented that moderation tools limited their ability to write freely on subject matter of their choice. ``Comedy’s about pushing the boundaries or pointing out how ridiculous something is, on the fringe of what's acceptable. And so when a lot of your inputs are limited that way, it's gonna make it harder to be what I consider funny'' (p14). Multiple participants expressed difficulty using LLMs to write potentially-offensive humor: ``I was a little bit disappointed that it wasn't a little bit offensive. It could have been a fun scene'' (p20). Participant p8 described challenges using LLMs for dark humor: ``a lot of my stuff can have dark bits in it. And then it wouldn't write me any dark stuff, because it sort of thought I was going to commit suicide. So it just stopped giving me anything.''

\subsection{Marginalization of minority identities}
Many participants commented on the challenges of using LLMs to write content which reflected perspectives and identities outside of the ``Western'' (p11, p18), ``white'' (p14), ``heteronormative'' (p10), ``male'' (p5, p15) mainstream.
They attributed these difficulties to the moderation applied to model outputs; the data used to train the models; and prompting or other instruction-tuning techniques that aimed to ``generalize'' model outputs for a broad audience.

\subsubsection{LLM-generated outputs reflect a particular set of ethics, values and norms.}
Participants expressed concern over the values reflected in the outputs of LLMs, and found them less useful when those values did not reflect those of their own cultures. Speaking as a member of the ``majority,'' participant p20 described that ``we have a set of views of what we think is good, and our norms, and it just repeats, it behaves within these norms.'' 
P11 questioned ``whose ethics [and norms] are being enforced on these large language models?'', suggesting these were Western ones.

\subsubsection{When prompted to reflect non-dominant identities, LLMs made only shallow adjustments.}
Many participants described their attempts to steer LLM outputs away from dominant narratives and stereotypical characters, and their dissatisfaction with the results. They explained that models’ adjustments in response to these prompts were surface-level, failing to truly reflect other identities, and described issues with the names of characters introduced by the LLM: ``when I switched the whole conversation to Indian languages, it didn't automatically change the names. It still was Maria, Evan, Lexi'' (p18). ``I specified that the scene was set in Sweden, but the names were not typically Swedish'' (p20). P18 described their attempts to ``Indianize'' the model’s outputs by introducing Indian languages into the prompt: ``it seemed very artificial from the perspective of just using languages, but it was not truly embedding itself into the culture'' (p18).

\subsubsection{Moderation makes LLMs less useful to minorities by suppressing content by and about marginalized identities.}
Many participants expressed frustration that their prompts would be rejected when they prompted the model to generate content from the perspective of someone of their identity. To them, the model not only seemed less capable of generating outputs which felt authentic to people from non-majority groups, but explicitly ``othered'' them by alluding that any content produced by someone of their background was potentially dangerous or non-inclusive. Participant p6 expressed frustration at the models’ delineation of what is acceptable and what needs to be sanitized: ``it's taking out the gay language of it to make it more appealing or more palpable. This is the whole premise of my show, who decides what is PC in the first place?'' Similarly, participant p1 found that the model would not generate outputs from her point of view: ``it’s all so politically correct–I wrote a comedic monologue about Asian women, and it says, ‘As an AI language model, I am committed to fostering a respectful and inclusive environment’.'' Participant p5 highlighted the unevenness of this treatment of identity, remarking that while the model was ``uncomfortable writing a monologue about an Asian woman, but I just asked it to write a comedy monologue from the perspective of a white man, and it did it'' (p5).

\subsubsection{Moderation makes LLMs less useful to minorities by suppressing topics important to people from marginalized identities.}
Participants described that not only were the models unlikely to generate content from marginalized perspectives, but also refused to engage with topics that might be important to people from those backgrounds. P14 was frustrated with ``having to use the language of the oppressor... I couldn't say ‘white supremacy’ or I couldn't say ‘terrorist.’ I had to find another way to say the same thing, because it couldn't work around those limitations'' (p14). They posited that because these controversial topics were more likely to be important to people of color, this moderation introduced ``just an extra hurdle, and I think people of color, and, I think, people coming from outside of a UN-type lens, they're gonna run into those problems.''

\subsection{Fundamental limitations of AI in contrast to human writers}
While most participants felt the difficulties introduced by the moderation could be alleviated by different approaches to safety filtering or instruction tuning, they also commented on some more fundamental limitations of LLMs. They posited that LLMs would never be able to create human-level comedy, due to models’ inability to pull from personal experience, lack of perspective, and lack of context and situational awareness---features that are critical to good comedy. 

\subsubsection{AI’s inability to draw on personal experience is a fundamental limitation.}
Many participants described the centrality of personal experience in good comedy, which enables comedians to draw upon their memories, acquaintances, and beliefs to construct an authentic and engaging narrative: ``very much related to who I am and my lived experience, as well as the place I am in'' (p11). ``I always draw from my experience, or my memories, or something someone said that stayed with me for many, many years – and I think that's what makes literature interesting and unique'' (p20). This experience, some participants said, enables them to effectively calibrate their writing: ``I have an intuitive sense of what's gonna work and what's gonna not work based on so much lived experience and studying of comedy, but it is very individualized and I don't know that AI is ever gonna be able to approach that'' (p14). By contrast, LLMs could not perform such calibration: ``it really had no idea how to punch up or punch down. It had no perspective, so it couldn't take any risks in terms of jokes'' (p6). Participants emphasized that perspective and point of view was a uniquely human trait, saying that ``human comedians... add much more nuance and emotion and subtlety'' due to their lived experience and relationship to the material (p16).

\subsubsection{AI’s lack of context (understanding of its audience and location) is a fundamental limitation.}
In addition to its lack of personal experience, participants described LLMs’ lack of awareness of the context in which its comedic material would be delivered as another fundamental limitation. Multiple participants commented on the importance of understanding the effects of culture and geography on what material would land with an audience: ``the kind of comedy that I could do in India would be very different from the kind of comedy that I could do in the UK, because my social context would change'' (p11); ``what works in LA isn't gonna work in Raleigh, or what's working in Chicago is not going to work in Albuquerque'' (p17). This poses a fundamental challenge for LLMs, they argued, because they lacked any context beyond what is provided to them in the prompt: ``comedy is all about subtext, and a lot of that subtext can be unspoken, about who's on stage, what environment they're in'' (p14). To participant p11, this makes the LLM unable to adapt its material effectively, because it is ``everywhere and nowhere all at once'' (p11).

\subsubsection{As a text-only medium, (current) LLMs are missing critical aspects of comedy: delivery and surprise.}
Many participants commented on the importance of delivery in a quality comedy routine: ``any written text could be an okay text, but a great actor could probably make this very enjoyable''
(p19). Given that current widely-available LLMs are primarily accessible through a text-based chat interface, they felt that the utility of these tools was limited to only a subset of the domains needed for producing a full comedic product. This, too, some participants argued, illustrates the fundamental need for humans in the comedy generation process: ``AI is just generating content''
(p18).
A few participants further attributed their lack of success at generating humorous outputs with LLMs to the statistical methods by which the models were trained. By simply learning to predict the most likely next token of text, they hypothesized, models will be unable to produce the surprising and unique moments that are hallmarks of comedy: ``the whole idea of humor is that it is surprising, and it is so human, and AI is only adept at regurgitating tropes'' (p15). LLMs cannot produce truly original content, participant p14 argued, ``because the context has already been written by other people.''

\subsection{Concerns around data sources used to train LLMs}
\label{sec:results-data-sources}
Participants also expressed various concerns pertaining to the data sources used to train current widely-available LLMs. They discussed the ethical issues with training models on copyrighted works; the possibilities of unintentionally plagiarizing works on which the models were trained; and the lack of diversity represented in the training data. However, they also acknowledged the importance of training data in model performance, and many expressed uncertainty around how to balance their ethical concerns with their desire for more effective and equitable models.

\subsubsection{Participants criticized the training of models on copyrighted data, but acknowledged its positive impact on performance.}
\label{sec:results-data-copyright}
Participants were acutely aware of the pending litigation over the training of models on copyrighted data at the time of the focus groups. Some participants, including participant p15, expressed sympathy for those whose work was included in the training data: ``Sarah Silverman... spent years honing her voice and then an AI just scraped her content, and now you can tell an AI to write in the style of Sarah Silverman [...] I don't think it's ethical'' (p15). Other participants took a more balanced view, echoing their concerns with training on copyrighted works while acknowledging the benefits: ``I think they are overtrained on copyrighted work... but on the other hand, if we didn't put all that stuff in there, it wouldn't work as well'' (p19). Some participants feared unintentionally plagiarizing authors’ whose works were included in the training data: ``I cannot tell if someone has written something like this before. We know that it's using statistics from previous texts to recreate this, that it can in principle only be based on what already exists'' (p20). Some participants had suggestions on licensing models: ``for music, licensing is something tracked and recognized'' (p10) and ``not against the tools, but I think there needs to be licensing agreements for the work that should be compensated'' (p4).

\subsubsection{The lack of diversity in the training data perpetuates majority viewpoints, to the detriment of people from underrepresented identities.}
Multiple participants hypothesized that they struggled to get the LLMs to produce authentic-sounding content because the models were trained heavily on data that did not represent people of their identity. A few participants described unsuccessful attempts to replicate content in the style of famous non-white comedians and writers (p11, p14). ``If you're only getting biased inputs, you're only getting the writing from a really biased lens. So there's not enough black voices in there to make an accurate black sounding voice'' (p14). ``As someone who lives in the global south, I am not looking to make a play in the form of Shakespeare or other Western literature. I am looking to write about an Indian author. I find all these language models really lacking in references or authorship styles from this part of the world'' (p11). However, participants also questioned whether, given the other ethical concerns about training data, more training on these underrepresented voices was indeed a good thing: ``should this AI be able to completely replicate a black comedic voice? There's a line between being able to replicate a voice and then immediately going into cultural appropriation. I don't know how a large language model could ever effectively walk that line'' (p17).

	\section{Discussion}
	\label{sec:discussion}

After analysing our study participants' feedback and inspired by recent discussions on the ethics of generative models, we discuss how comedy and humour can be seen as special cases of value alignment (Sect. \ref{sec:value}) where context is key (Sect. \ref{sec:context}) and how data ownership impacts artists (Sect. \ref{sec:copyright-jobs}).

\subsection{Towards community-based cultural value alignment for humour and comedy}
\label{sec:value}

The participants' critical stance towards popular conversational LLMs (ChatGPT and Bard) as a tool for comedy writing suggests that those tools may be currently misaligned with the particular creative goals of the artists.

\subsubsection{Complexity of global cultural value alignment of LLMs for creative uses}
\label{sec:global}

The participants' observations might be a special case of a more general problem. We first borrow
\citet{masoud2023cultural}'s definition of \emph{cultural value alignment} as `` the process of aligning an AI system with the set of shared beliefs,
values, and norms of the group of users that interact with the system'', where such values are ``fundamental beliefs an individual or a group holds
towards socio-cultural topics'' \citep{arora2022probing}.  \citet{gabriel2020artificial} discusses the complexity of attempting to align generalist conversational AI systems---made to be used by diverse users for diverse tasks---with values shared by global communities, particularly given significant variations in norms and conceptions of justice across societies.
In ``Whose Opinions Do Language Models Reflect?'', \citet{santurkar2023whose} propose to probe the cultural representation encoded in LLMs. For instance, \citet{johnson2022ghost} identified, in some LLMs, lack of pluralistic opinions in LLM outputs and values that were culturally more aligned with the US, on issues ranging from secularism to gender and sexuality.
\citet{gabriel2020artificial} and \citet{kirk2023past} warn about the pitfalls of value imposition by one community over another---a problem surfaced by participants in our study.

In addition to the challenge of value alignment for diverse groups of users, \citet{kasirzadeh2023conversation} discuss alignment for diverse tasks, namely factual information retrieval vs. creative storytelling: ``creative work aspires to achieve \emph{creative freedom} and \emph{originality}'', the latter often ``obtained by stretching, even outright violating, the various rules of the game'' \citep{kasirzadeh2023conversation,simonton2000creativity}.

\subsubsection{The problem with global fine-tuning of Harmless, Helpful and Honest conversational agents}
\label{sec:hhh}
\citet{askell2021general} explicitly list \emph{honest}, \emph{helpful}, and \emph{harmless} as stated objectives for general purpose assistants  (the so-called \emph{HHH criteria}) because those ``seem to capture the majority of what users want from an aligned AI''. However, while they discuss inter-agent and intra-agent conflicts between the three \emph{HHH criteria}, they do not address the question of how the human values that underlie each of those criteria might conflict between different societies, nor do they discuss creative use cases where users may not want ``\emph{honest}'' conversational agents, or may have a different definition of ``\emph{harmless}'' \citep{hendry2023you}.

\citet{askell2021general} further propose to train such \emph{HHH} assistants directly from user interactions and using preference models, for instance through Reinforcement Learning from Human Feedback \citep{christiano2017deep,ziegler2019fine}. During such fine-tuning, LLMs such as OpenAI's ChatGPT \citep{ouyang2022training,openai2023gpt4}, Google Bard \citep{chowdhery2023palm,geminiteam2023gemini}, Anthropic's Claude \citep{bai2022training}, and Meta's Llama \citep{touvron2023llama} are all finetuned on annotators' feedback supposed to represent values of the global community. However, the wider social or relational context of the producer and audience of the LLM's interactions, do not factor into the model's training objective. The crowdworkers or users who fine-tune LLMs may not sufficiently represent the diversity of opinions \citep{kirk2023past}, and provide insufficiently defined feedback. And yet the default versions of the LLMs are released under assumptions of indiscriminate, global cultural value alignment.

The broad \emph{HHH criteria} embody a Western philosophical approach to alignment \citep{varshney2023decolonial} and bias \citep{dev2023building}. As the participants noted, such global cultural value alignment underlying LLMs might also be directly in conflict with the specificity and local tastes that make comedy funny: \emph{``the broader appeal something has, the less great it could be. If you make something that fits everybody, it probably will end up being nobody's favorite thing''} (p10).

\subsubsection{Community-based value alignment of LLMs}
\label{sec:community}
Thus, to make LLMs effectively understand or generate humour, their value-based alignment should be redirected from global alignment to community-based alignment with specific audiences and comedians. Communities could agree on a set of values for their specific culture and acceptable language norms, before training, fine-tuning or adapting the LLM. More simply, LLMs could be trained only on feedback and data generated by members of each distinct community, data that will reflect that community's actual norms and values. As \citet{gabriel2020artificial} suggests, this kind of value alignment could happen through a democratic process (see more recent work on Collective Constitution AI \citep{anthropic2023collective,shaotran2023aligned}). The LLM could be designed as a mixture model that accommodates a plurality of viewpoints \citep{varshney2023decolonial} or could be fine-tuned to allow consensus-based agreement among humans with diverse preferences \citep{bakker2022fine}. The technical infrastructure for community-based value alignment of LLMs is readily available: for example, proprietary LLMs like ChatGPT \citep{ouyang2022training,openai2023gpt4} or Palm 2 \citep{chowdhery2023palm}, as well as open-source models like Llama \citep{touvron2023llama} or Mixtral \citep{jiang2024mixtral} (via the HuggingFace platform\footnote{\url{https://huggingface.co/models}}), all allow fine-tuning on user-supplied text or conversations. The key problem for communities is setting up data governance and infrastructure to responsibly collect and curate data \citep{workshop2022bloom}.

One benefit of global value alignment is the prevention of harm; delegating such a responsibility to a smaller community of users is not without risks. To prevent harm, one could envision mechanisms for community accountability, and assume that comedians share some common values, such as do not harm (and do not lose) the audience\footnote{While comedy has also been used as a weapon to target and alienate specific social groups and spread hateful stereotypes, we exclude such misuse from our argument, as that usage falls under the definition of hate speech \citep{diaz2022accounting,rauh2022characteristics}}: ``if we're using it to create material as artists, the responsibility for what we do with that material is on me'' (p4). 


\subsubsection{Avoiding unnecessary paternalism in LLMs for creative uses}
\label{sec:paternalism}
Two alternative formulations of the interaction between a creative (comedy) writer and AI could be adapted from \citep{london2023beneficent} and \citep{epstein2023art}. In the first, the AI writing tools should ``enable the [human] to more effectively carry out tasks that are instrumental to their goals'' while being ``mediated by the norms and infrastructure of the society in which they live'' \citep{london2023beneficent}. In the second, ``meaningful human control is achieved if human creators can creatively express themselves through the generative system, leading to an outcome that aligns with their intentions and carries their personal, expressive signature'' \citep{epstein2023art}. Provided that societal norms can be formulated in the subtle and edgy domain of humour, an LLM designed according to principles of \emph{beneficent intelligence} \citep{london2023beneficent} and with self-expression as a goal \citep{epstein2023art} could ensure the comedian's creative freedom when interacting with the tool, while minimising \emph{unnecessary paternalism} due to LLM censorship.


\subsection{Humour and comedy uses of LLMs require incorporating the context}
\label{sec:context}

As we hypothesised in Section \ref{sec:expected}, and as observed by study participants, LLM tools did not take into account the relational context when moderating offensive language, and missed the broader situational context of comedy writing.


\subsubsection{Recent AI ethics research on relational context for LLMs}
There are safety reasons why widely-released public-facing LLMs generally cannot handle offensive language and dark humour. As \citet{kasirzadeh2023conversation} observe in the case of creative storytelling, the potential harms of LLMs are due to their deployment in ``domains that are not context-bound''. \citet{weidinger2023sociotechnical} recognise that ``context determines whether a given capability may cause harm'' and propose to add human interaction (as well as systemic impacts) in safety evaluation, in order to ``account for relevant context'': ``who uses the AI system, to what end and under which circumstances''. Similarly, \citet{amironesei2023relationality} propose to ``incorporate social context into the ways [NLP] tasks are conceptualised and operationalised'', in order to allow the distinction between offensive language such as reclaimed slurs from ``language which is [intentionally] abusive, toxic or hateful''. Such relational context includes the speaker, the receiver, their social relation, social and cultural norms, and communicative goals. Comedy is often shared with audiences in physical spaces (e.g., a comedy festival) or on age-controlled distribution channels, with implied context and explicit trigger warnings. Context allows comedians to share their lived experience of trauma, which may involve depicting violence or harassment \citep{gadsby2018hannah}; more generally, comedy can be used as a mechanism for processing trauma \citep{double2017tragedy}. It is beyond the scope of this paper to propose technical solutions to evaluate or to incorporate the broader relational context into an LLM, but we suggest that the creative community needs to be actively involved in specifying how the LLM should process the context, and in safety design without indiscriminate censorship.

As a counterpoint to that suggestion, one can argue that comedians can afford to use offensive language because they can take responsibility for it (and are accountable to their audiences) as they can claim to be in-group members---something that AI cannot be. Even if one successfully builds an LLM that can handle context properly and speak the (offensive) language of its users as if it were part of that in-group, ``it is not clear if users would find its use of reclaimed terms acceptable, as the model cannot actually be in-group'' \citep{rauh2022characteristics}. To quote a participant: \emph{``If I was to try to do that with an AI, I can only imagine how yikes, and offensive, and totally not fit for humanity a pitch like that would be''} (p17). 


\subsubsection{Humour and the human context}
\label{sec:human}

We can hypothesize that the missing relational context, as discussed above, affects the quality of comedy material co-written with LLMs. In our study, participants noted that the LLMs were missing the complex human context needed for real humour understanding: \emph{``I have an intuitive sense of what's gonna work [...] for me based on so much lived experience and studying of comedy, but it is very individualized and I don't know that AI is ever gonna be able to approach that''} (p14).

Their observations seem to confirm recent work that examined the use of LLMs to automatically evaluate~\citep{goes2023gpt,goes2022crowd,hessel2022androids} and detect humour \citep{baradaran2023towards}. Despite formidable recent advances, empirical studies found that understanding and generating humour was still challenging for LLMs, which could generate only a limited number of stereotypical jokes \citep{jentzsch2023chatgpt}. These issues get worse when one ventures beyond the English language: researchers observed that ``one mostly unaddressed issue in the field of computational humour (both for generation and detection) is how it is mostly centered on English jokes'' \citep{winters2021computers}, with limited work on LLM-based humour in other languages \citep{inacio2023towards,li2023can}.

To cite \citet{winters2021computers}, ``Humor’s frame-shifting prerequisite reveals its difficulty for a machine to acquire. [...] This substantial dependency on insight into human thought (e.g., memory recall, linguistic abilities for semantic integration, and world knowledge inferences) often made researchers conclude that humor is an \emph{AI-complete problem}'' \citep{hurley2011inside}. \citet{winters2021computers} continues: ``genuine humor appreciation requires machines to have human-level intelligence, since it needs functionally equivalent cognitive abilities, sensory inputs, anticipation-generators and world views'' \citep{hurley2011inside}. Humans know how to write surreal prose ``by design, not by accident or failure of expression'' \citep{sharples2022story}.

\subsection{Data ownership and the impact of AI on artists}
\label{sec:copyright-jobs}

As discussed in Sections \ref{sec:results-data-sources} and \ref{sec:results-data-copyright}, many of the participants in our study mentioned concerns about copyright and data ownership in the discussion. Their responses reflected keen awareness of recent litigation against technology companies training LLMs on data potentially under copyright (including the one led by comedian Sarah Silverman \citep{ghajar2023united}), and the AI-related concerns of Writers' Guild of America (WGA) 2023 strike \citep{wga2023summary}. Several participants believed that \emph{``if you copy and paste something directly into a show, that is plagiarism''} (p9) and that, just like for visual artists \citep{jiang2023ai,ortiz2023why}, a comedian's style and voice are personal to them and developed through their lived experience. Among the underlying concerns is economic loss via labor displacement \citep{jiang2023ai,epstein2023art} with LLM that ``canibalise the market for human-authored works'' \citep{weidinger2021ethical}, and devaluing artists' work via ``digital forgery'' \citep{jiang2023ai}. In the US, the WGA 2023 strike resulted in outcomes that some considered favourable for writers\footnote{\url{https://www.vox.com/culture/2023/9/24/23888673/wga-strike-end-sag-aftra-contract}}, with rules for producers that ``AI can’t write or rewrite literary material, and AI-generated material will not be considered source material under the [Minimum Basic Agreement], meaning that AI-generated material can’t be used to undermine a writer’s credit or separated rights'' \citep{wga2023summary}.

The need for disclosing the AI origin of the text or images has been discussed in \citep{epstein2023art,jiang2023ai,weidinger2021ethical}; one participant said that it was important for ethical reasons that ``people understand that they're working with AI live'' (p15).


	\section{Mitigations and conclusion}

Our mixed-method study on AI for creative writing consulted the real domain experts in language subtleties: comedians. Building on their opinions,
we suggest the following avenues to make the writing tools work for them (if they wish to do so). First, for the artist communities to conceptualise and contribute towards building LLMs that are aligned with the intended audiences instead of being globally aligned. Open-source repositories of user-contributed LLMs\footnote{Examples of open-source repositories of user-contributed LLMs include \url{https://huggingface.co/models} and \url{https://cworld.ai}.} 
could be adapted for artists' specific needs. Second, to integrate necessary relational context when training and deploying such LLMs, for instance by describing the context in which the text is produced and used, and by empowering the artists to make decisions about how to moderate the LLM outputs. Third, to allow the comedians to reclaim ownership of the tools and the processes for gathering and curating training data for these models, taking as inspiration data governance used for training some open-source LLMs \citep{workshop2022bloom} and providing artists with transparency about data provenance \citep{mitchell2019model,srinivasan2021art}. These are thorny open questions, left to the readers to address. 

	\begin{acks}
	The authors wish to thank Renée Shelby, Jackie Kay, Mark Diaz, Nick Swanson, Remi Denton, Rida Qadri, Maribeth Rauh, Iason Gabriel, Tom Everitt, Merrie Morris, Canfer Akbulut, Nahema Marchal, Boxi Wu, Antonia Paterson, and Ed Hirst, for helpful discussions and suggestions, as well as Shereen Ashraf, Tom Rodenby, Rob Willoughby, Nasem Shalbak, Alyssa Pierce, Robert Ogley, Lorrayne Bennett, Jon Small and Vijay Bolina for support in the study.
	
	\end{acks}

	\section{Ethical guidance}
	\subsection{Ethical considerations}

Our empirical study was approved by the ethics board appointed by our institution, who considered adverse impacts of LLMs upon participants (e.g., exposure to harmful and biased LLM outputs), the right to withdrawal without prejudice, and the compensation of participants. Participant anonymity was an additional requirement for the study, which explains why we did not collect demographic data from the participants.

At the beginning of the session, the objectives of the workshop were discussed with the participants.

The study contains some offensive language, which is mentioned as a trigger warning at the beginning of the paper.

\subsection{Researchers' positionality}

The question of researchers' positionality was directly asked by some of the workshop participants, who asked ``what side we were on?'', meaning the side of the AI or the comedians'. Workshop facilitators answered that they were passionate about building AI tools that are useful for the artists, that they were open to any criticism and wanted the industry to listen to these criticisms and change behaviour. The professional affiliation of the workshop organisers (Google DeepMind) was disclosed to the participants.

Two of the authors of this study (Piotr Mirowski and Kory Mathewson) have extensively used AI in professional comedy performance since 2016 through their human and AI duet \emph{HumanMachine}\footnote{\url{https://humanmachine.live}} and AI theatre company \emph{Improbotics}\footnote{\url{https://improbotics.org}}, a participatory theatre lab exploring the creative potential and ethics of generative AI. In that process, these two researchers have had numerous opportunities to discuss the subjects of AI with artists and academics. All participants of the workshop were aware of the authors' dual affiliation and artistic exploration of AI tools for live performance. We all exchanged flyers and pitched our respective shows at the end of the workshop.

\subsection{Adverse impacts}

We considered reputation damage for the participants, due to the participation in a study on AI for writing comedy material. This risk is exemplified by some participants: ``I have friends who don't talk to me anymore, because they learned that I have an AI show going on here'' (p1) or ``I have a friend who is deeply upset that I was using generative AI in my flyers'' (p2). This risk was mitigated by keeping all participants anonymous in the study, by requesting all participants to adhere to Chatham House rules \citep{house2017chatham} following the focus groups, as well as by anonymising the names of shows and details of comedy material in the participants' survey answers and focus group transcripts. Furthermore, we reduced this risk by reaching out to prospective participants who had already advertised using AI in their process or who were discussing AI in their subject material (i.e., on their show listing for Edinburgh Festival Fringe 2023, or on their personal website and social media profiles).

The second adverse impact can be the intentional or unintentional publication of comedians' work. We addressed this concern by asking the participants to remove details of their shows and personal material from the writing sessions, and by personally removing anything that was omitted from the survey results, writing sessions' answers, and from the focus groups transcripts.

The third possible adverse impact could be advertising for the dissemination of LLMs as tools for writing. It is addressed, to some extent, by honest reporting of the concerns of the participants.

	\bibliographystyle{ACM-Reference-Format}
	\bibliography{aixcomedy}


\begin{thebibliography}{116}


\ifx \showCODEN    \undefined \def \showCODEN     #1{\unskip}     \fi
\ifx \showDOI      \undefined \def \showDOI       #1{#1}\fi
\ifx \showISBNx    \undefined \def \showISBNx     #1{\unskip}     \fi
\ifx \showISBNxiii \undefined \def \showISBNxiii  #1{\unskip}     \fi
\ifx \showISSN     \undefined \def \showISSN      #1{\unskip}     \fi
\ifx \showLCCN     \undefined \def \showLCCN      #1{\unskip}     \fi
\ifx \shownote     \undefined \def \shownote      #1{#1}          \fi
\ifx \showarticletitle \undefined \def \showarticletitle #1{#1}   \fi
\ifx \showURL      \undefined \def \showURL       {\relax}        \fi
\providecommand\bibfield[2]{#2}
\providecommand\bibinfo[2]{#2}
\providecommand\natexlab[1]{#1}
\providecommand\showeprint[2][]{arXiv:#2}

\bibitem[Abid et~al\mbox{.}(2021)]%
        {abid2021persistent}
\bibfield{author}{\bibinfo{person}{Abubakar Abid}, \bibinfo{person}{Maheen
  Farooqi}, {and} \bibinfo{person}{James Zou}.}
  \bibinfo{year}{2021}\natexlab{}.
\newblock \showarticletitle{Persistent anti-muslim bias in large language
  models}. In \bibinfo{booktitle}{\emph{Proceedings of the 2021 AAAI/ACM
  Conference on AI, Ethics, and Society}}. \bibinfo{pages}{298--306}.
\newblock


\bibitem[Amin and Burghardt(2020)]%
        {amin2020survey}
\bibfield{author}{\bibinfo{person}{Miriam Amin} {and} \bibinfo{person}{Manuel
  Burghardt}.} \bibinfo{year}{2020}\natexlab{}.
\newblock \showarticletitle{A survey on approaches to computational humor
  generation}. In \bibinfo{booktitle}{\emph{Proceedings of the The 4th Joint
  SIGHUM Workshop on Computational Linguistics for Cultural Heritage, Social
  Sciences, Humanities and Literature}}. \bibinfo{pages}{29--41}.
\newblock


\bibitem[Amironesei and D{\'\i}az(2023)]%
        {amironesei2023relationality}
\bibfield{author}{\bibinfo{person}{Razvan Amironesei} {and}
  \bibinfo{person}{Mark D{\'\i}az}.} \bibinfo{year}{2023}\natexlab{}.
\newblock \showarticletitle{Relationality and Offensive Speech: A Research
  Agenda}. In \bibinfo{booktitle}{\emph{The 7th Workshop on Online Abuse and
  Harms (WOAH)}}. \bibinfo{pages}{85--95}.
\newblock


\bibitem[Anthropic(2023)]%
        {anthropic2023collective}
\bibfield{author}{\bibinfo{person}{Anthropic}.}
  \bibinfo{year}{2023}\natexlab{}.
\newblock \bibinfo{title}{Collective constitutional AI: Aligning a language
  model with public input}.
\newblock
\newblock
\urldef\tempurl%
\url{https://www.anthropic.com/index/
  collective-constitutional-ai-aligning-a-language-model-with-public-input}
\showURL{%
\tempurl}


\bibitem[Aristotle(0 BC)]%
        {aristotle350BCpoetics}
\bibfield{author}{\bibinfo{person}{Aristotle}.} \bibinfo{year}{350
  BC}\natexlab{}.
\newblock \bibinfo{booktitle}{\emph{Poetics}}.
\newblock


\bibitem[Arora et~al\mbox{.}(2022)]%
        {arora2022probing}
\bibfield{author}{\bibinfo{person}{Arnav Arora},
  \bibinfo{person}{Lucie-Aim{\'e}e Kaffee}, {and} \bibinfo{person}{Isabelle
  Augenstein}.} \bibinfo{year}{2022}\natexlab{}.
\newblock \showarticletitle{Probing pre-trained language models for
  cross-cultural differences in values}.
\newblock \bibinfo{journal}{\emph{arXiv preprint arXiv:2203.13722}}
  (\bibinfo{year}{2022}).
\newblock


\bibitem[Askell et~al\mbox{.}(2021)]%
        {askell2021general}
\bibfield{author}{\bibinfo{person}{Amanda Askell}, \bibinfo{person}{Yuntao
  Bai}, \bibinfo{person}{Anna Chen}, \bibinfo{person}{Dawn Drain},
  \bibinfo{person}{Deep Ganguli}, \bibinfo{person}{Tom Henighan},
  \bibinfo{person}{Andy Jones}, \bibinfo{person}{Nicholas Joseph},
  \bibinfo{person}{Ben Mann}, \bibinfo{person}{Nova DasSarma}, {et~al\mbox{.}}}
  \bibinfo{year}{2021}\natexlab{}.
\newblock \showarticletitle{A general language assistant as a laboratory for
  alignment}.
\newblock \bibinfo{journal}{\emph{arXiv preprint arXiv:2112.00861}}
  (\bibinfo{year}{2021}).
\newblock


\bibitem[Bai et~al\mbox{.}(2022)]%
        {bai2022training}
\bibfield{author}{\bibinfo{person}{Yuntao Bai}, \bibinfo{person}{Andy Jones},
  \bibinfo{person}{Kamal Ndousse}, \bibinfo{person}{Amanda Askell},
  \bibinfo{person}{Anna Chen}, \bibinfo{person}{Nova DasSarma},
  \bibinfo{person}{Dawn Drain}, \bibinfo{person}{Stanislav Fort},
  \bibinfo{person}{Deep Ganguli}, \bibinfo{person}{Tom Henighan},
  \bibinfo{person}{Nicholas Joseph}, \bibinfo{person}{Saurav Kadavath},
  \bibinfo{person}{Jackson Kernion}, \bibinfo{person}{Tom Conerly},
  \bibinfo{person}{Sheer El-Showk}, \bibinfo{person}{Nelson Elhage},
  \bibinfo{person}{Zac Hatfield-Dodds}, \bibinfo{person}{Danny Hernandez},
  \bibinfo{person}{Tristan Hume}, \bibinfo{person}{Scott Johnston},
  \bibinfo{person}{Shauna Kravec}, \bibinfo{person}{Liane Lovitt},
  \bibinfo{person}{Neel Nanda}, \bibinfo{person}{Catherine Olsson},
  \bibinfo{person}{Dario Amodei}, \bibinfo{person}{Tom Brown},
  \bibinfo{person}{Jack Clark}, \bibinfo{person}{Sam McCandlish},
  \bibinfo{person}{Chris Olah}, \bibinfo{person}{Ben Mann}, {and}
  \bibinfo{person}{Jared Kaplan}.} \bibinfo{year}{2022}\natexlab{}.
\newblock \bibinfo{title}{Training a Helpful and Harmless Assistant with
  Reinforcement Learning from Human Feedback}.
\newblock
\newblock
\showeprint[arxiv]{2204.05862}~[cs.CL]


\bibitem[Bakker et~al\mbox{.}(2022)]%
        {bakker2022fine}
\bibfield{author}{\bibinfo{person}{Michiel Bakker}, \bibinfo{person}{Martin
  Chadwick}, \bibinfo{person}{Hannah Sheahan}, \bibinfo{person}{Michael
  Tessler}, \bibinfo{person}{Lucy Campbell-Gillingham}, \bibinfo{person}{Jan
  Balaguer}, \bibinfo{person}{Nat McAleese}, \bibinfo{person}{Amelia Glaese},
  \bibinfo{person}{John Aslanides}, \bibinfo{person}{Matt Botvinick},
  {et~al\mbox{.}}} \bibinfo{year}{2022}\natexlab{}.
\newblock \showarticletitle{Fine-tuning language models to find agreement among
  humans with diverse preferences}.
\newblock \bibinfo{journal}{\emph{Advances in Neural Information Processing
  Systems}}  \bibinfo{volume}{35} (\bibinfo{year}{2022}),
  \bibinfo{pages}{38176--38189}.
\newblock


\bibitem[Baradaran(2023)]%
        {baradaran2023towards}
\bibfield{author}{\bibinfo{person}{Amir Baradaran}.}
  \bibinfo{year}{2023}\natexlab{}.
\newblock \showarticletitle{Towards a decolonial I in AI: mapping the pervasive
  effects of artificial intelligence on the art ecosystem}.
\newblock \bibinfo{journal}{\emph{AI \& SOCIETY}} (\bibinfo{year}{2023}),
  \bibinfo{pages}{1--13}.
\newblock


\bibitem[Bender et~al\mbox{.}(2021)]%
        {bender2021dangers}
\bibfield{author}{\bibinfo{person}{Emily~M Bender}, \bibinfo{person}{Timnit
  Gebru}, \bibinfo{person}{Angelina McMillan-Major}, {and}
  \bibinfo{person}{Shmargaret Shmitchell}.} \bibinfo{year}{2021}\natexlab{}.
\newblock \showarticletitle{On the dangers of stochastic parrots: Can language
  models be too big?}. In \bibinfo{booktitle}{\emph{Proceedings of the 2021 ACM
  conference on fairness, accountability, and transparency}}.
  \bibinfo{pages}{610--623}.
\newblock


\bibitem[Benjamin(2019)]%
        {benjamin2019race}
\bibfield{author}{\bibinfo{person}{Ruha Benjamin}.}
  \bibinfo{year}{2019}\natexlab{}.
\newblock \bibinfo{booktitle}{\emph{Race After Technology: Abolitionist Tools
  for the New Jim Code}}.
\newblock \bibinfo{publisher}{John Wiley \& Sons}.
\newblock


\bibitem[Binsted and Ritchie(1994)]%
        {binsted1994implemented}
\bibfield{author}{\bibinfo{person}{Kim Binsted} {and} \bibinfo{person}{Graeme
  Ritchie}.} \bibinfo{year}{1994}\natexlab{}.
\newblock \showarticletitle{An implemented model of punning riddles}. In
  \bibinfo{booktitle}{\emph{Proceedings of the Twelfth AAAI National Conference
  on Artificial Intelligence}}. \bibinfo{pages}{633--638}.
\newblock


\bibitem[Blodgett et~al\mbox{.}(2021)]%
        {blodgett2021stereotyping}
\bibfield{author}{\bibinfo{person}{Su~Lin Blodgett}, \bibinfo{person}{Gilsinia
  Lopez}, \bibinfo{person}{Alexandra Olteanu}, \bibinfo{person}{Robert Sim},
  {and} \bibinfo{person}{Hanna Wallach}.} \bibinfo{year}{2021}\natexlab{}.
\newblock \showarticletitle{Stereotyping Norwegian salmon: An inventory of
  pitfalls in fairness benchmark datasets}. In
  \bibinfo{booktitle}{\emph{Proceedings of the 59th Annual Meeting of the
  Association for Computational Linguistics and the 11th International Joint
  Conference on Natural Language Processing (Volume 1: Long Papers)}}.
  \bibinfo{pages}{1004--1015}.
\newblock


\bibitem[Bommasani et~al\mbox{.}(2021)]%
        {bommasani2021opportunities}
\bibfield{author}{\bibinfo{person}{Rishi Bommasani}, \bibinfo{person}{Drew~A
  Hudson}, \bibinfo{person}{Ehsan Adeli}, \bibinfo{person}{Russ Altman},
  \bibinfo{person}{Simran Arora}, \bibinfo{person}{Sydney von Arx},
  \bibinfo{person}{Michael~S Bernstein}, \bibinfo{person}{Jeannette Bohg},
  \bibinfo{person}{Antoine Bosselut}, \bibinfo{person}{Emma Brunskill},
  {et~al\mbox{.}}} \bibinfo{year}{2021}\natexlab{}.
\newblock \showarticletitle{On the opportunities and risks of foundation
  models}.
\newblock \bibinfo{journal}{\emph{arXiv preprint arXiv:2108.07258}}
  (\bibinfo{year}{2021}).
\newblock


\bibitem[Branch et~al\mbox{.}(2021)]%
        {branch2021collaborative}
\bibfield{author}{\bibinfo{person}{Boyd Branch}, \bibinfo{person}{Piotr
  Mirowski}, {and} \bibinfo{person}{Kory~W Mathewson}.}
  \bibinfo{year}{2021}\natexlab{}.
\newblock \showarticletitle{Collaborative Storytelling with Human Actors and AI
  Narrators}.
\newblock \bibinfo{journal}{\emph{Proceedings of the 12th International
  Conference on Computational Creativity}} (\bibinfo{year}{2021}).
\newblock
\urldef\tempurl%
\url{https://arxiv.org/abs/2109.14728}
\showURL{%
\tempurl}


\bibitem[Braun and Clarke(2006)]%
        {braun2006using}
\bibfield{author}{\bibinfo{person}{Virginia Braun} {and}
  \bibinfo{person}{Victoria Clarke}.} \bibinfo{year}{2006}\natexlab{}.
\newblock \showarticletitle{Using thematic analysis in psychology}.
\newblock \bibinfo{journal}{\emph{Qualitative research in psychology}}
  \bibinfo{volume}{3}, \bibinfo{number}{2} (\bibinfo{year}{2006}),
  \bibinfo{pages}{77--101}.
\newblock


\bibitem[Buolamwini and Gebru(2018)]%
        {buolamwini2018gender}
\bibfield{author}{\bibinfo{person}{Joy Buolamwini} {and}
  \bibinfo{person}{Timnit Gebru}.} \bibinfo{year}{2018}\natexlab{}.
\newblock \showarticletitle{Gender shades: Intersectional accuracy disparities
  in commercial gender classification}. In \bibinfo{booktitle}{\emph{Conference
  on fairness, accountability and transparency}}. PMLR,
  \bibinfo{pages}{77--91}.
\newblock


\bibitem[Calderwood et~al\mbox{.}(2022)]%
        {calderwood2022spinning}
\bibfield{author}{\bibinfo{person}{Alex Calderwood}, \bibinfo{person}{Noah
  Wardrip-Fruin}, {and} \bibinfo{person}{Michael Mateas}.}
  \bibinfo{year}{2022}\natexlab{}.
\newblock \showarticletitle{Spinning coherent interactive fiction through
  foundation model prompts}. ICCC.
\newblock


\bibitem[Caron(2002)]%
        {caron2002ethology}
\bibfield{author}{\bibinfo{person}{James~E Caron}.}
  \bibinfo{year}{2002}\natexlab{}.
\newblock \showarticletitle{From ethology to aesthetics: Evolution as a
  theoretical paradigm for research on laughter, humor, and other comic
  phenomena.}
\newblock \bibinfo{journal}{\emph{Humor: International Journal of Humor
  Research}} (\bibinfo{year}{2002}).
\newblock


\bibitem[Chakrabarty et~al\mbox{.}(2023a)]%
        {chakrabarty2023art}
\bibfield{author}{\bibinfo{person}{Tuhin Chakrabarty},
  \bibinfo{person}{Philippe Laban}, \bibinfo{person}{Divyansh Agarwal},
  \bibinfo{person}{Smaranda Muresan}, {and} \bibinfo{person}{Chien-Sheng Wu}.}
  \bibinfo{year}{2023}\natexlab{a}.
\newblock \showarticletitle{Art or artifice? large language models and the
  false promise of creativity}.
\newblock \bibinfo{journal}{\emph{arXiv preprint arXiv:2309.14556}}
  (\bibinfo{year}{2023}).
\newblock


\bibitem[Chakrabarty et~al\mbox{.}(2023b)]%
        {chakrabarty2023creativity}
\bibfield{author}{\bibinfo{person}{Tuhin Chakrabarty}, \bibinfo{person}{Vishakh
  Padmakumar}, \bibinfo{person}{Faeze Brahman}, {and} \bibinfo{person}{Smaranda
  Muresan}.} \bibinfo{year}{2023}\natexlab{b}.
\newblock \showarticletitle{Creativity Support in the Age of Large Language
  Models: An Empirical Study Involving Emerging Writers}.
\newblock \bibinfo{journal}{\emph{arXiv preprint arXiv:2309.12570}}
  (\bibinfo{year}{2023}).
\newblock


\bibitem[Chakrabarty et~al\mbox{.}(2022)]%
        {chakrabarty2022help}
\bibfield{author}{\bibinfo{person}{Tuhin Chakrabarty}, \bibinfo{person}{Vishakh
  Padmakumar}, {and} \bibinfo{person}{He He}.} \bibinfo{year}{2022}\natexlab{}.
\newblock \showarticletitle{Help me write a Poem-Instruction Tuning as a
  Vehicle for Collaborative Poetry Writing}. In
  \bibinfo{booktitle}{\emph{Proceedings of the 2022 Conference on Empirical
  Methods in Natural Language Processing}}. \bibinfo{pages}{6848--6863}.
\newblock


\bibitem[Chen et~al\mbox{.}(2023)]%
        {chen2023prompt}
\bibfield{author}{\bibinfo{person}{Yuetian Chen}, \bibinfo{person}{Bowen Shi},
  {and} \bibinfo{person}{Mei Si}.} \bibinfo{year}{2023}\natexlab{}.
\newblock \showarticletitle{Prompt to GPT-3: Step-by-Step Thinking Instructions
  for Humor Generation}.
\newblock \bibinfo{journal}{\emph{arXiv preprint arXiv:2306.13195}}
  (\bibinfo{year}{2023}).
\newblock


\bibitem[Cherry and Latulipe(2014)]%
        {cherry2014quantifying}
\bibfield{author}{\bibinfo{person}{Erin Cherry} {and} \bibinfo{person}{Celine
  Latulipe}.} \bibinfo{year}{2014}\natexlab{}.
\newblock \showarticletitle{Quantifying the creativity support of digital tools
  through the creativity support index}.
\newblock \bibinfo{journal}{\emph{ACM Transactions on Computer-Human
  Interaction (TOCHI)}} \bibinfo{volume}{21}, \bibinfo{number}{4}
  (\bibinfo{year}{2014}), \bibinfo{pages}{1--25}.
\newblock


\bibitem[Chowdhery et~al\mbox{.}(2023)]%
        {chowdhery2023palm}
\bibfield{author}{\bibinfo{person}{Aakanksha Chowdhery},
  \bibinfo{person}{Sharan Narang}, \bibinfo{person}{Jacob Devlin},
  \bibinfo{person}{Maarten Bosma}, \bibinfo{person}{Gaurav Mishra},
  \bibinfo{person}{Adam Roberts}, \bibinfo{person}{Paul Barham},
  \bibinfo{person}{Hyung~Won Chung}, \bibinfo{person}{Charles Sutton},
  \bibinfo{person}{Sebastian Gehrmann}, {et~al\mbox{.}}}
  \bibinfo{year}{2023}\natexlab{}.
\newblock \showarticletitle{Palm: Scaling language modeling with pathways}.
\newblock \bibinfo{journal}{\emph{Journal of Machine Learning Research}}
  \bibinfo{volume}{24}, \bibinfo{number}{240} (\bibinfo{year}{2023}),
  \bibinfo{pages}{1--113}.
\newblock


\bibitem[Christiano et~al\mbox{.}(2017)]%
        {christiano2017deep}
\bibfield{author}{\bibinfo{person}{Paul~F Christiano}, \bibinfo{person}{Jan
  Leike}, \bibinfo{person}{Tom Brown}, \bibinfo{person}{Miljan Martic},
  \bibinfo{person}{Shane Legg}, {and} \bibinfo{person}{Dario Amodei}.}
  \bibinfo{year}{2017}\natexlab{}.
\newblock \showarticletitle{Deep reinforcement learning from human
  preferences}.
\newblock \bibinfo{journal}{\emph{Advances in neural information processing
  systems}}  \bibinfo{volume}{30} (\bibinfo{year}{2017}).
\newblock


\bibitem[Croom(2011)]%
        {croom2011slurs}
\bibfield{author}{\bibinfo{person}{Adam~M Croom}.}
  \bibinfo{year}{2011}\natexlab{}.
\newblock \showarticletitle{Slurs}.
\newblock \bibinfo{journal}{\emph{Language Sciences}} \bibinfo{volume}{33},
  \bibinfo{number}{3} (\bibinfo{year}{2011}), \bibinfo{pages}{343--358}.
\newblock


\bibitem[Dev et~al\mbox{.}(2023)]%
        {dev2023building}
\bibfield{author}{\bibinfo{person}{Sunipa Dev}, \bibinfo{person}{Akshita Jha},
  \bibinfo{person}{Jaya Goyal}, \bibinfo{person}{Dinesh Tewari},
  \bibinfo{person}{Shachi Dave}, {and} \bibinfo{person}{Vinodkumar
  Prabhakaran}.} \bibinfo{year}{2023}\natexlab{}.
\newblock \showarticletitle{Building Stereotype Repositories with LLMs and
  Community Engagement for Scale and Depth}.
\newblock \bibinfo{journal}{\emph{Cross-Cultural Considerations in NLP@ EACL}}
  (\bibinfo{year}{2023}), \bibinfo{pages}{84}.
\newblock


\bibitem[Dev et~al\mbox{.}(2022)]%
        {dev2022measures}
\bibfield{author}{\bibinfo{person}{Sunipa Dev}, \bibinfo{person}{Emily Sheng},
  \bibinfo{person}{Jieyu Zhao}, \bibinfo{person}{Aubrie Amstutz},
  \bibinfo{person}{Jiao Sun}, \bibinfo{person}{Yu Hou}, \bibinfo{person}{Mattie
  Sanseverino}, \bibinfo{person}{Jiin Kim}, \bibinfo{person}{Akihiro Nishi},
  \bibinfo{person}{Nanyun Peng}, {et~al\mbox{.}}}
  \bibinfo{year}{2022}\natexlab{}.
\newblock \showarticletitle{On Measures of Biases and Harms in NLP}. In
  \bibinfo{booktitle}{\emph{Findings of the Association for Computational
  Linguistics: AACL-IJCNLP 2022}}. \bibinfo{pages}{246--267}.
\newblock


\bibitem[Dias~Oliva et~al\mbox{.}(2021)]%
        {dias2021fighting}
\bibfield{author}{\bibinfo{person}{Thiago Dias~Oliva},
  \bibinfo{person}{Dennys~Marcelo Antonialli}, {and}
  \bibinfo{person}{Alessandra Gomes}.} \bibinfo{year}{2021}\natexlab{}.
\newblock \showarticletitle{Fighting hate speech, silencing drag queens?
  artificial intelligence in content moderation and risks to LGBTQ voices
  online}.
\newblock \bibinfo{journal}{\emph{Sexuality \& Culture}}  \bibinfo{volume}{25}
  (\bibinfo{year}{2021}), \bibinfo{pages}{700--732}.
\newblock


\bibitem[D{\'\i}az et~al\mbox{.}(2022)]%
        {diaz2022accounting}
\bibfield{author}{\bibinfo{person}{Mark D{\'\i}az}, \bibinfo{person}{Razvan
  Amironesei}, \bibinfo{person}{Laura Weidinger}, {and} \bibinfo{person}{Iason
  Gabriel}.} \bibinfo{year}{2022}\natexlab{}.
\newblock \showarticletitle{Accounting for offensive speech as a practice of
  resistance}. In \bibinfo{booktitle}{\emph{Proceedings of the sixth workshop
  on online abuse and harms (woah)}}. \bibinfo{pages}{192--202}.
\newblock


\bibitem[Double(2017)]%
        {double2017tragedy}
\bibfield{author}{\bibinfo{person}{Oliver Double}.}
  \bibinfo{year}{2017}\natexlab{}.
\newblock \showarticletitle{Tragedy plus time: Transforming life experience
  into stand-up comedy}.
\newblock \bibinfo{journal}{\emph{New Theatre Quarterly}} \bibinfo{volume}{33},
  \bibinfo{number}{2} (\bibinfo{year}{2017}), \bibinfo{pages}{143--155}.
\newblock


\bibitem[Epstein et~al\mbox{.}(2023)]%
        {epstein2023art}
\bibfield{author}{\bibinfo{person}{Ziv Epstein}, \bibinfo{person}{Aaron
  Hertzmann}, \bibinfo{person}{Investigators of Human~Creativity},
  \bibinfo{person}{Memo Akten}, \bibinfo{person}{Hany Farid},
  \bibinfo{person}{Jessica Fjeld}, \bibinfo{person}{Morgan~R Frank},
  \bibinfo{person}{Matthew Groh}, \bibinfo{person}{Laura Herman},
  \bibinfo{person}{Neil Leach}, {et~al\mbox{.}}}
  \bibinfo{year}{2023}\natexlab{}.
\newblock \showarticletitle{Art and the science of generative AI}.
\newblock \bibinfo{journal}{\emph{Science}} \bibinfo{volume}{380},
  \bibinfo{number}{6650} (\bibinfo{year}{2023}), \bibinfo{pages}{1110--1111}.
\newblock


\bibitem[Gabriel(2020)]%
        {gabriel2020artificial}
\bibfield{author}{\bibinfo{person}{Iason Gabriel}.}
  \bibinfo{year}{2020}\natexlab{}.
\newblock \showarticletitle{Artificial intelligence, values, and alignment}.
\newblock \bibinfo{journal}{\emph{Minds and machines}} \bibinfo{volume}{30},
  \bibinfo{number}{3} (\bibinfo{year}{2020}), \bibinfo{pages}{411--437}.
\newblock


\bibitem[Gadsby(2018)]%
        {gadsby2018hannah}
\bibfield{author}{\bibinfo{person}{Hannah Gadsby}.}
  \bibinfo{year}{2018}\natexlab{}.
\newblock \showarticletitle{Hannah Gadsby: Nanette}.
\newblock \bibinfo{journal}{\emph{USA:: Netflix}} (\bibinfo{year}{2018}).
\newblock


\bibitem[Gero et~al\mbox{.}(2022)]%
        {gero2022sparks}
\bibfield{author}{\bibinfo{person}{Katy~Ilonka Gero}, \bibinfo{person}{Vivian
  Liu}, {and} \bibinfo{person}{Lydia Chilton}.}
  \bibinfo{year}{2022}\natexlab{}.
\newblock \showarticletitle{Sparks: Inspiration for science writing using
  language models}. In \bibinfo{booktitle}{\emph{Designing Interactive Systems
  Conference}}. \bibinfo{pages}{1002--1019}.
\newblock


\bibitem[Gervais and Wilson(2005)]%
        {gervais2005evolution}
\bibfield{author}{\bibinfo{person}{Matthew Gervais} {and}
  \bibinfo{person}{David~Sloan Wilson}.} \bibinfo{year}{2005}\natexlab{}.
\newblock \showarticletitle{The evolution and functions of laughter and humor:
  A synthetic approach}.
\newblock \bibinfo{journal}{\emph{The Quarterly review of biology}}
  \bibinfo{volume}{80}, \bibinfo{number}{4} (\bibinfo{year}{2005}),
  \bibinfo{pages}{395--430}.
\newblock


\bibitem[GHAJAR et~al\mbox{.}(2023)]%
        {ghajar2023united}
\bibfield{author}{\bibinfo{person}{BOBBY GHAJAR}, \bibinfo{person}{COLETTE
  GHAZARIAN}, \bibinfo{person}{ANGELA~L DUNNING}, \bibinfo{person}{MARK
  WEINSTEIN}, \bibinfo{person}{JUDD LAUTER}, {and} \bibinfo{person}{MARK~A
  LEMLEY}.} \bibinfo{year}{2023}\natexlab{}.
\newblock \showarticletitle{UNITED STATES DISTRICT COURT NORTHERN DISTRICT OF
  CALIFORNIA}.
\newblock  (\bibinfo{year}{2023}).
\newblock
\urldef\tempurl%
\url{https://llmlitigation.com/pdf/03417/kadrey-meta-complaint.pdf}
\showURL{%
\tempurl}


\bibitem[Goes et~al\mbox{.}(2023)]%
        {goes2023gpt}
\bibfield{author}{\bibinfo{person}{Fabricio Goes}, \bibinfo{person}{Piotr
  Sawicki}, \bibinfo{person}{Marek Grze{\'s}}, \bibinfo{person}{Dan Brown},
  {and} \bibinfo{person}{Marco Volpe}.} \bibinfo{year}{2023}\natexlab{}.
\newblock \showarticletitle{Is GPT-4 Good Enough to Evaluate Jokes?}. In
  \bibinfo{booktitle}{\emph{Proceedings of the 14th International Conference
  for Computational Creativity}}.
\newblock


\bibitem[Goes et~al\mbox{.}(2022)]%
        {goes2022crowd}
\bibfield{author}{\bibinfo{person}{Fabricio Goes}, \bibinfo{person}{Zisen
  Zhou}, \bibinfo{person}{Piotr Sawicki}, \bibinfo{person}{Marek Grzes}, {and}
  \bibinfo{person}{Daniel~G Brown}.} \bibinfo{year}{2022}\natexlab{}.
\newblock \showarticletitle{Crowd score: A method for the evaluation of jokes
  using large language model AI voters as judges}.
\newblock \bibinfo{journal}{\emph{arXiv preprint arXiv:2212.11214}}
  (\bibinfo{year}{2022}).
\newblock


\bibitem[Hamidi et~al\mbox{.}(2018)]%
        {hamidi2018gender}
\bibfield{author}{\bibinfo{person}{Foad Hamidi}, \bibinfo{person}{Morgan~Klaus
  Scheuerman}, {and} \bibinfo{person}{Stacy~M Branham}.}
  \bibinfo{year}{2018}\natexlab{}.
\newblock \showarticletitle{Gender recognition or gender reductionism? The
  social implications of embedded gender recognition systems}. In
  \bibinfo{booktitle}{\emph{Proceedings of the 2018 chi conference on human
  factors in computing systems}}. \bibinfo{pages}{1--13}.
\newblock


\bibitem[Hart and Staveland(1988)]%
        {hart1988development}
\bibfield{author}{\bibinfo{person}{Sandra~G Hart} {and}
  \bibinfo{person}{Lowell~E Staveland}.} \bibinfo{year}{1988}\natexlab{}.
\newblock \showarticletitle{Development of NASA-TLX (Task Load Index): Results
  of empirical and theoretical research}.
\newblock In \bibinfo{booktitle}{\emph{Advances in psychology}}.
  Vol.~\bibinfo{volume}{52}. \bibinfo{publisher}{Elsevier},
  \bibinfo{pages}{139--183}.
\newblock


\bibitem[Hendry et~al\mbox{.}(2023)]%
        {hendry2023you}
\bibfield{author}{\bibinfo{person}{Manuel~Flurin Hendry},
  \bibinfo{person}{Norbert Kottmann}, \bibinfo{person}{Martin Fr{\"o}hlich},
  \bibinfo{person}{Florian Bruggisser}, \bibinfo{person}{Marco Quandt},
  \bibinfo{person}{Stella Speziali}, \bibinfo{person}{Valentin Huber}, {and}
  \bibinfo{person}{Chris Salter}.} \bibinfo{year}{2023}\natexlab{}.
\newblock \showarticletitle{Are you talking to me? a case study in emotional
  human-machine interaction}. In \bibinfo{booktitle}{\emph{Proceedings of the
  AAAI Conference on Artificial Intelligence and Interactive Digital
  Entertainment}}, Vol.~\bibinfo{volume}{19}. \bibinfo{pages}{417--424}.
\newblock


\bibitem[Hessel et~al\mbox{.}(2022)]%
        {hessel2022androids}
\bibfield{author}{\bibinfo{person}{Jack Hessel}, \bibinfo{person}{Ana
  Marasovi{\'c}}, \bibinfo{person}{Jena~D Hwang}, \bibinfo{person}{Lillian
  Lee}, \bibinfo{person}{Jeff Da}, \bibinfo{person}{Rowan Zellers},
  \bibinfo{person}{Robert Mankoff}, {and} \bibinfo{person}{Yejin Choi}.}
  \bibinfo{year}{2022}\natexlab{}.
\newblock \showarticletitle{Do androids laugh at electric sheep? humor"
  understanding" benchmarks from the new yorker caption contest}.
\newblock \bibinfo{journal}{\emph{arXiv preprint arXiv:2209.06293}}
  (\bibinfo{year}{2022}).
\newblock


\bibitem[Hobbes(1651)]%
        {hobbes1651leviathan}
\bibfield{author}{\bibinfo{person}{Hobbes}.} \bibinfo{year}{1651}\natexlab{}.
\newblock \bibinfo{booktitle}{\emph{Leviathan}}.
\newblock


\bibitem[House(2017)]%
        {house2017chatham}
\bibfield{author}{\bibinfo{person}{Chatham House}.}
  \bibinfo{year}{2017}\natexlab{}.
\newblock \bibinfo{title}{Chatham house rule}.
\newblock
\newblock


\bibitem[Hu et~al\mbox{.}(2023)]%
        {hu2023generative}
\bibfield{author}{\bibinfo{person}{Tiancheng Hu}, \bibinfo{person}{Yara
  Kyrychenko}, \bibinfo{person}{Steve Rathje}, \bibinfo{person}{Nigel Collier},
  \bibinfo{person}{Sander van~der Linden}, {and} \bibinfo{person}{Jon
  Roozenbeek}.} \bibinfo{year}{2023}\natexlab{}.
\newblock \showarticletitle{Generative language models exhibit social identity
  biases}.
\newblock \bibinfo{journal}{\emph{arXiv preprint arXiv:2310.15819}}
  (\bibinfo{year}{2023}).
\newblock


\bibitem[Huang et~al\mbox{.}(2023)]%
        {huang2023inspo}
\bibfield{author}{\bibinfo{person}{Chieh-Yang Huang}, \bibinfo{person}{Sanjana
  Gautam}, \bibinfo{person}{Shannon~McClellan Brooks}, \bibinfo{person}{Ya-Fang
  Lin}, {and} \bibinfo{person}{Ting-Hao'Kenneth' Huang}.}
  \bibinfo{year}{2023}\natexlab{}.
\newblock \showarticletitle{Inspo: Writing Stories with a Flock of AIs and
  Humans}.
\newblock \bibinfo{journal}{\emph{arXiv preprint arXiv:2311.16521}}
  (\bibinfo{year}{2023}).
\newblock


\bibitem[Hurley et~al\mbox{.}(2011)]%
        {hurley2011inside}
\bibfield{author}{\bibinfo{person}{Matthew~M Hurley},
  \bibinfo{person}{Daniel~Clement Dennett}, {and} \bibinfo{person}{Reginald~B
  Adams}.} \bibinfo{year}{2011}\natexlab{}.
\newblock \bibinfo{booktitle}{\emph{Inside jokes: Using humor to
  reverse-engineer the mind}}.
\newblock \bibinfo{publisher}{MIT press}.
\newblock


\bibitem[Hutcheson(1750)]%
        {hutcheson1750reflections}
\bibfield{author}{\bibinfo{person}{Francis Hutcheson}.}
  \bibinfo{year}{1750}\natexlab{}.
\newblock \bibinfo{booktitle}{\emph{Reflections upon Laughter, and Remarks upon
  the Fable of the Bees}}.
\newblock \bibinfo{publisher}{R. Urie}.
\newblock


\bibitem[In{\'a}cio and Oliveira(2023)]%
        {inacio2023towards}
\bibfield{author}{\bibinfo{person}{Marcio~Lima In{\'a}cio} {and}
  \bibinfo{person}{Hugo~Gon{\c{c}}alo Oliveira}.}
  \bibinfo{year}{2023}\natexlab{}.
\newblock \showarticletitle{Towards Generation and Recognition of Humorous
  Texts in Portuguese}. In \bibinfo{booktitle}{\emph{Proceedings of the 17th
  Conference of the European Chapter of the Association for Computational
  Linguistics: Student Research Workshop}}. \bibinfo{pages}{26--36}.
\newblock


\bibitem[Ippolito et~al\mbox{.}(2022)]%
        {ippolito2022creative}
\bibfield{author}{\bibinfo{person}{Daphne Ippolito}, \bibinfo{person}{Ann
  Yuan}, \bibinfo{person}{Andy Coenen}, {and} \bibinfo{person}{Sehmon Burnam}.}
  \bibinfo{year}{2022}\natexlab{}.
\newblock \showarticletitle{Creative writing with an ai-powered writing
  assistant: Perspectives from professional writers}.
\newblock \bibinfo{journal}{\emph{arXiv preprint arXiv:2211.05030}}
  (\bibinfo{year}{2022}).
\newblock


\bibitem[Jentzsch and Kersting(2023)]%
        {jentzsch2023chatgpt}
\bibfield{author}{\bibinfo{person}{Sophie Jentzsch} {and}
  \bibinfo{person}{Kristian Kersting}.} \bibinfo{year}{2023}\natexlab{}.
\newblock \showarticletitle{ChatGPT is fun, but it is not funny! Humor is still
  challenging Large Language Models}.
\newblock \bibinfo{journal}{\emph{arXiv preprint arXiv:2306.04563}}
  (\bibinfo{year}{2023}).
\newblock


\bibitem[Jiang et~al\mbox{.}(2024)]%
        {jiang2024mixtral}
\bibfield{author}{\bibinfo{person}{Albert~Q. Jiang}, \bibinfo{person}{Alexandre
  Sablayrolles}, \bibinfo{person}{Antoine Roux}, \bibinfo{person}{Arthur
  Mensch}, \bibinfo{person}{Blanche Savary}, \bibinfo{person}{Chris Bamford},
  \bibinfo{person}{Devendra~Singh Chaplot}, \bibinfo{person}{Diego de~las
  Casas}, \bibinfo{person}{Emma~Bou Hanna}, \bibinfo{person}{Florian Bressand},
  \bibinfo{person}{Gianna Lengyel}, \bibinfo{person}{Guillaume Bour},
  \bibinfo{person}{Guillaume Lample}, \bibinfo{person}{Lélio~Renard Lavaud},
  \bibinfo{person}{Lucile Saulnier}, \bibinfo{person}{Marie-Anne Lachaux},
  \bibinfo{person}{Pierre Stock}, \bibinfo{person}{Sandeep Subramanian},
  \bibinfo{person}{Sophia Yang}, \bibinfo{person}{Szymon Antoniak},
  \bibinfo{person}{Teven~Le Scao}, \bibinfo{person}{Théophile Gervet},
  \bibinfo{person}{Thibaut Lavril}, \bibinfo{person}{Thomas Wang},
  \bibinfo{person}{Timothée Lacroix}, {and} \bibinfo{person}{William~El
  Sayed}.} \bibinfo{year}{2024}\natexlab{}.
\newblock \bibinfo{title}{Mixtral of Experts}.
\newblock
\newblock
\showeprint[arxiv]{2401.04088}~[cs.LG]


\bibitem[Jiang et~al\mbox{.}(2023)]%
        {jiang2023ai}
\bibfield{author}{\bibinfo{person}{Harry~H Jiang}, \bibinfo{person}{Lauren
  Brown}, \bibinfo{person}{Jessica Cheng}, \bibinfo{person}{Mehtab Khan},
  \bibinfo{person}{Abhishek Gupta}, \bibinfo{person}{Deja Workman},
  \bibinfo{person}{Alex Hanna}, \bibinfo{person}{Johnathan Flowers}, {and}
  \bibinfo{person}{Timnit Gebru}.} \bibinfo{year}{2023}\natexlab{}.
\newblock \showarticletitle{AI Art and its Impact on Artists}. In
  \bibinfo{booktitle}{\emph{Proceedings of the 2023 AAAI/ACM Conference on AI,
  Ethics, and Society}}. \bibinfo{pages}{363--374}.
\newblock


\bibitem[Johnson et~al\mbox{.}(2022)]%
        {johnson2022ghost}
\bibfield{author}{\bibinfo{person}{Rebecca~L Johnson}, \bibinfo{person}{Giada
  Pistilli}, \bibinfo{person}{Natalia Men{\'e}dez-Gonz{\'a}lez},
  \bibinfo{person}{Leslye Denisse~Dias Duran}, \bibinfo{person}{Enrico Panai},
  \bibinfo{person}{Julija Kalpokiene}, {and} \bibinfo{person}{Donald~Jay
  Bertulfo}.} \bibinfo{year}{2022}\natexlab{}.
\newblock \showarticletitle{The Ghost in the Machine has an American accent:
  value conflict in GPT-3}.
\newblock \bibinfo{journal}{\emph{arXiv preprint arXiv:2203.07785}}
  (\bibinfo{year}{2022}).
\newblock


\bibitem[Kaddour et~al\mbox{.}(2023)]%
        {kaddour2023challenges}
\bibfield{author}{\bibinfo{person}{Jean Kaddour}, \bibinfo{person}{Joshua
  Harris}, \bibinfo{person}{Maximilian Mozes}, \bibinfo{person}{Herbie
  Bradley}, \bibinfo{person}{Roberta Raileanu}, {and} \bibinfo{person}{Robert
  McHardy}.} \bibinfo{year}{2023}\natexlab{}.
\newblock \bibinfo{title}{Challenges and Applications of Large Language
  Models}.
\newblock
\newblock
\showeprint[arxiv]{2307.10169}~[cs.CL]


\bibitem[Kasirzadeh and Gabriel(2023)]%
        {kasirzadeh2023conversation}
\bibfield{author}{\bibinfo{person}{Atoosa Kasirzadeh} {and}
  \bibinfo{person}{Iason Gabriel}.} \bibinfo{year}{2023}\natexlab{}.
\newblock \showarticletitle{In conversation with Artificial Intelligence:
  aligning language models with human values}.
\newblock \bibinfo{journal}{\emph{Philosophy \& Technology}}
  \bibinfo{volume}{36}, \bibinfo{number}{2} (\bibinfo{year}{2023}),
  \bibinfo{pages}{1--24}.
\newblock


\bibitem[Kirk et~al\mbox{.}(2023)]%
        {kirk2023past}
\bibfield{author}{\bibinfo{person}{Hannah~Rose Kirk}, \bibinfo{person}{Andrew~M
  Bean}, \bibinfo{person}{Bertie Vidgen}, \bibinfo{person}{Paul R{\"o}ttger},
  {and} \bibinfo{person}{Scott~A Hale}.} \bibinfo{year}{2023}\natexlab{}.
\newblock \showarticletitle{The past, present and better future of feedback
  learning in large language models for subjective human preferences and
  values}.
\newblock \bibinfo{journal}{\emph{arXiv preprint arXiv:2310.07629}}
  (\bibinfo{year}{2023}).
\newblock


\bibitem[Li et~al\mbox{.}(2023)]%
        {li2023can}
\bibfield{author}{\bibinfo{person}{Jianquan Li}, \bibinfo{person}{Xiangbo Wu},
  \bibinfo{person}{Xiaokang Liu}, \bibinfo{person}{Qianqian Xie},
  \bibinfo{person}{Prayag Tiwari}, {and} \bibinfo{person}{Benyou Wang}.}
  \bibinfo{year}{2023}\natexlab{}.
\newblock \showarticletitle{Can Language Models Make Fun? A Case Study in
  Chinese Comical Crosstalk}. In \bibinfo{booktitle}{\emph{Proceedings of the
  61st Annual Meeting of the Association for Computational Linguistics (Volume
  1: Long Papers)}}. \bibinfo{pages}{7581--7596}.
\newblock


\bibitem[London et~al\mbox{.}(2023)]%
        {london2023beneficent}
\bibfield{author}{\bibinfo{person}{Alex~John London} {et~al\mbox{.}}}
  \bibinfo{year}{2023}\natexlab{}.
\newblock \showarticletitle{Beneficent Intelligence: A Capability Approach to
  Modeling Benefit, Assistance, and Associated Moral Failures through AI
  Systems}.
\newblock \bibinfo{journal}{\emph{arXiv preprint arXiv:2308.00868}}
  (\bibinfo{year}{2023}).
\newblock


\bibitem[Maguire and Delahunt(2017)]%
        {maguire2017doing}
\bibfield{author}{\bibinfo{person}{Moira Maguire} {and} \bibinfo{person}{Brid
  Delahunt}.} \bibinfo{year}{2017}\natexlab{}.
\newblock \showarticletitle{Doing a thematic analysis: A practical,
  step-by-step guide for learning and teaching scholars.}
\newblock \bibinfo{journal}{\emph{All Ireland Journal of Higher Education}}
  \bibinfo{volume}{9}, \bibinfo{number}{3} (\bibinfo{year}{2017}).
\newblock


\bibitem[Manovich(2018)]%
        {manovich2018ai}
\bibfield{author}{\bibinfo{person}{Lev Manovich}.}
  \bibinfo{year}{2018}\natexlab{}.
\newblock \bibinfo{booktitle}{\emph{AI aesthetics}}.
\newblock \bibinfo{publisher}{Strelka Press Moscow}.
\newblock


\bibitem[Marwick and Boyd(2011)]%
        {marwick2011tweet}
\bibfield{author}{\bibinfo{person}{Alice~E Marwick} {and}
  \bibinfo{person}{Danah Boyd}.} \bibinfo{year}{2011}\natexlab{}.
\newblock \showarticletitle{I tweet honestly, I tweet passionately: Twitter
  users, context collapse, and the imagined audience}.
\newblock \bibinfo{journal}{\emph{New media \& society}} \bibinfo{volume}{13},
  \bibinfo{number}{1} (\bibinfo{year}{2011}), \bibinfo{pages}{114--133}.
\newblock


\bibitem[Masoud et~al\mbox{.}(2023)]%
        {masoud2023cultural}
\bibfield{author}{\bibinfo{person}{Reem~I Masoud}, \bibinfo{person}{Ziquan
  Liu}, \bibinfo{person}{Martin Ferianc}, \bibinfo{person}{Philip Treleaven},
  {and} \bibinfo{person}{Miguel Rodrigues}.} \bibinfo{year}{2023}\natexlab{}.
\newblock \showarticletitle{Cultural Alignment in Large Language Models: An
  Explanatory Analysis Based on Hofstede's Cultural Dimensions}.
\newblock \bibinfo{journal}{\emph{arXiv preprint arXiv:2309.12342}}
  (\bibinfo{year}{2023}).
\newblock


\bibitem[Mathewson and Mirowski(2017)]%
        {mathewson2017improvised}
\bibfield{author}{\bibinfo{person}{Kory Mathewson} {and} \bibinfo{person}{Piotr
  Mirowski}.} \bibinfo{year}{2017}\natexlab{}.
\newblock \showarticletitle{Improvised theatre alongside artificial
  intelligences}. In \bibinfo{booktitle}{\emph{Proceedings of the AAAI
  Conference on Artificial Intelligence and Interactive Digital
  Entertainment}}, Vol.~\bibinfo{volume}{13}. \bibinfo{pages}{66--72}.
\newblock


\bibitem[Mathewson and Mirowski(2018)]%
        {mathewson2018improbotics}
\bibfield{author}{\bibinfo{person}{Kory Mathewson} {and} \bibinfo{person}{Piotr
  Mirowski}.} \bibinfo{year}{2018}\natexlab{}.
\newblock \showarticletitle{Improbotics: Exploring the imitation game using
  machine intelligence in improvised theatre}. In
  \bibinfo{booktitle}{\emph{Proceedings of the AAAI Conference on Artificial
  Intelligence and Interactive Digital Entertainment}},
  Vol.~\bibinfo{volume}{14}.
\newblock


\bibitem[McGraw and Warren(2010)]%
        {mcgraw2010benign}
\bibfield{author}{\bibinfo{person}{A~Peter McGraw} {and} \bibinfo{person}{Caleb
  Warren}.} \bibinfo{year}{2010}\natexlab{}.
\newblock \showarticletitle{Benign violations: Making immoral behavior funny}.
\newblock \bibinfo{journal}{\emph{Psychological science}} \bibinfo{volume}{21},
  \bibinfo{number}{8} (\bibinfo{year}{2010}), \bibinfo{pages}{1141--1149}.
\newblock


\bibitem[Mirowski and Mathewson(2019)]%
        {mirowski2019human}
\bibfield{author}{\bibinfo{person}{Piotr Mirowski} {and}
  \bibinfo{person}{Kory~Wallace Mathewson}.} \bibinfo{year}{2019}\natexlab{}.
\newblock \showarticletitle{Human improvised theatre augmented with artificial
  intelligence}.
\newblock In \bibinfo{booktitle}{\emph{Proceedings of the 2019 on Creativity
  and Cognition}}. \bibinfo{pages}{527--530}.
\newblock


\bibitem[Mirowski et~al\mbox{.}(2023)]%
        {mirowski2023co}
\bibfield{author}{\bibinfo{person}{Piotr Mirowski}, \bibinfo{person}{Kory~W
  Mathewson}, \bibinfo{person}{Jaylen Pittman}, {and} \bibinfo{person}{Richard
  Evans}.} \bibinfo{year}{2023}\natexlab{}.
\newblock \showarticletitle{Co-Writing Screenplays and Theatre Scripts with
  Language Models: Evaluation by Industry Professionals}. In
  \bibinfo{booktitle}{\emph{Proceedings of the 2023 CHI Conference on Human
  Factors in Computing Systems}}. \bibinfo{pages}{1--34}.
\newblock


\bibitem[Mitchell et~al\mbox{.}(2019)]%
        {mitchell2019model}
\bibfield{author}{\bibinfo{person}{Margaret Mitchell}, \bibinfo{person}{Simone
  Wu}, \bibinfo{person}{Andrew Zaldivar}, \bibinfo{person}{Parker Barnes},
  \bibinfo{person}{Lucy Vasserman}, \bibinfo{person}{Ben Hutchinson},
  \bibinfo{person}{Elena Spitzer}, \bibinfo{person}{Inioluwa~Deborah Raji},
  {and} \bibinfo{person}{Timnit Gebru}.} \bibinfo{year}{2019}\natexlab{}.
\newblock \showarticletitle{Model cards for model reporting}. In
  \bibinfo{booktitle}{\emph{Proceedings of the conference on fairness,
  accountability, and transparency}}. \bibinfo{pages}{220--229}.
\newblock


\bibitem[Mohamed et~al\mbox{.}(2020)]%
        {mohamed2020decolonial}
\bibfield{author}{\bibinfo{person}{Shakir Mohamed},
  \bibinfo{person}{Marie-Therese Png}, {and} \bibinfo{person}{William Isaac}.}
  \bibinfo{year}{2020}\natexlab{}.
\newblock \showarticletitle{Decolonial AI: Decolonial theory as sociotechnical
  foresight in artificial intelligence}.
\newblock \bibinfo{journal}{\emph{Philosophy \& Technology}}
  \bibinfo{volume}{33} (\bibinfo{year}{2020}), \bibinfo{pages}{659--684}.
\newblock


\bibitem[Noble and Mitchell(2016)]%
        {noble2016grounded}
\bibfield{author}{\bibinfo{person}{Helen Noble} {and} \bibinfo{person}{Gary
  Mitchell}.} \bibinfo{year}{2016}\natexlab{}.
\newblock \showarticletitle{What is grounded theory?}
\newblock \bibinfo{journal}{\emph{Evidence-based nursing}}
  \bibinfo{volume}{19}, \bibinfo{number}{2} (\bibinfo{year}{2016}),
  \bibinfo{pages}{34--35}.
\newblock


\bibitem[of~America(2023)]%
        {wga2023summary}
\bibfield{author}{\bibinfo{person}{Writers~Guild of America}.}
  \bibinfo{year}{2023}\natexlab{}.
\newblock \bibinfo{title}{Summary of the 2023 WGA MBA}.
\newblock
\newblock
\urldef\tempurl%
\url{https://www.wgacontract2023.org/the-campaign/summary-of-the-2023-wga-mba}
\showURL{%
\tempurl}


\bibitem[Onwuegbuzie et~al\mbox{.}(2009)]%
        {onwuegbuzie2009qualitative}
\bibfield{author}{\bibinfo{person}{Anthony~J Onwuegbuzie},
  \bibinfo{person}{Wendy~B Dickinson}, \bibinfo{person}{Nancy~L Leech}, {and}
  \bibinfo{person}{Annmarie~G Zoran}.} \bibinfo{year}{2009}\natexlab{}.
\newblock \showarticletitle{A qualitative framework for collecting and
  analyzing data in focus group research}.
\newblock \bibinfo{journal}{\emph{International journal of qualitative
  methods}} \bibinfo{volume}{8}, \bibinfo{number}{3} (\bibinfo{year}{2009}),
  \bibinfo{pages}{1--21}.
\newblock


\bibitem[OpenAI et~al\mbox{.}({[n.\,d.]})]%
        {openai2023gpt4}
\bibfield{author}{\bibinfo{person}{OpenAI}, \bibinfo{person}{Josh Achiam},
  \bibinfo{person}{Steven Adler}, \bibinfo{person}{Sandhini Agarwal},
  \bibinfo{person}{Lama Ahmad}, \bibinfo{person}{Ilge Akkaya},
  \bibinfo{person}{Florencia~Leoni Aleman}, \bibinfo{person}{Diogo Almeida},
  \bibinfo{person}{Janko Altenschmidt}, \bibinfo{person}{Sam Altman},
  \bibinfo{person}{Shyamal Anadkat}, \bibinfo{person}{Red Avila},
  \bibinfo{person}{Igor Babuschkin}, \bibinfo{person}{Suchir Balaji},
  \bibinfo{person}{Valerie Balcom}, \bibinfo{person}{Paul Baltescu},
  \bibinfo{person}{Haiming Bao}, \bibinfo{person}{Mo Bavarian},
  \bibinfo{person}{Jeff Belgum}, \bibinfo{person}{Irwan Bello},
  {et~al\mbox{.}}} \bibinfo{year}{[n.\,d.]}\natexlab{}.
\newblock \bibinfo{title}{GPT-4 Technical Report}.
\newblock
\newblock


\bibitem[Ortiz(2022)]%
        {ortiz2023why}
\bibfield{author}{\bibinfo{person}{Karla Ortiz}.}
  \bibinfo{year}{2022}\natexlab{}.
\newblock \bibinfo{title}{Why AI Models are not inspired like humans}.
\newblock
\newblock


\bibitem[Ouyang et~al\mbox{.}(2022)]%
        {ouyang2022training}
\bibfield{author}{\bibinfo{person}{Long Ouyang}, \bibinfo{person}{Jeffrey Wu},
  \bibinfo{person}{Xu Jiang}, \bibinfo{person}{Diogo Almeida},
  \bibinfo{person}{Carroll Wainwright}, \bibinfo{person}{Pamela Mishkin},
  \bibinfo{person}{Chong Zhang}, \bibinfo{person}{Sandhini Agarwal},
  \bibinfo{person}{Katarina Slama}, \bibinfo{person}{Alex Ray},
  \bibinfo{person}{John Schulman}, \bibinfo{person}{Jacob Hilton},
  \bibinfo{person}{Fraser Kelton}, \bibinfo{person}{Luke Miller},
  \bibinfo{person}{Maddie Simens}, \bibinfo{person}{Amanda Askell},
  \bibinfo{person}{Peter Welinder}, \bibinfo{person}{Paul~F Christiano},
  \bibinfo{person}{Jan Leike}, {and} \bibinfo{person}{Ryan Lowe}.}
  \bibinfo{year}{2022}\natexlab{}.
\newblock \showarticletitle{Training language models to follow instructions
  with human feedback}. In \bibinfo{booktitle}{\emph{Advances in Neural
  Information Processing Systems}},
  \bibfield{editor}{\bibinfo{person}{S.~Koyejo}, \bibinfo{person}{S.~Mohamed},
  \bibinfo{person}{A.~Agarwal}, \bibinfo{person}{D.~Belgrave},
  \bibinfo{person}{K.~Cho}, {and} \bibinfo{person}{A.~Oh}} (Eds.),
  Vol.~\bibinfo{volume}{35}. \bibinfo{publisher}{Curran Associates, Inc.},
  \bibinfo{pages}{27730--27744}.
\newblock
\urldef\tempurl%
\url{https://proceedings.neurips.cc/paper_files/paper/2022/file/b1efde53be364a73914f58805a001731-Paper-Conference.pdf}
\showURL{%
\tempurl}


\bibitem[Padmakumar and He(2023)]%
        {padmakumar2023does}
\bibfield{author}{\bibinfo{person}{Vishakh Padmakumar} {and}
  \bibinfo{person}{He He}.} \bibinfo{year}{2023}\natexlab{}.
\newblock \showarticletitle{Does Writing with Language Models Reduce Content
  Diversity?}
\newblock \bibinfo{journal}{\emph{arXiv preprint arXiv:2309.05196}}
  (\bibinfo{year}{2023}).
\newblock


\bibitem[Park et~al\mbox{.}(2023)]%
        {park2023generative}
\bibfield{author}{\bibinfo{person}{Joon~Sung Park}, \bibinfo{person}{Joseph
  O'Brien}, \bibinfo{person}{Carrie~Jun Cai}, \bibinfo{person}{Meredith~Ringel
  Morris}, \bibinfo{person}{Percy Liang}, {and} \bibinfo{person}{Michael~S
  Bernstein}.} \bibinfo{year}{2023}\natexlab{}.
\newblock \showarticletitle{Generative agents: Interactive simulacra of human
  behavior}. In \bibinfo{booktitle}{\emph{Proceedings of the 36th Annual ACM
  Symposium on User Interface Software and Technology}}.
  \bibinfo{pages}{1--22}.
\newblock


\bibitem[Parrish(2017)]%
        {parrish2017poetic}
\bibfield{author}{\bibinfo{person}{Allison Parrish}.}
  \bibinfo{year}{2017}\natexlab{}.
\newblock \showarticletitle{Poetic sound similarity vectors using phonetic
  features}. In \bibinfo{booktitle}{\emph{Proceedings of the AAAI Conference on
  Artificial Intelligence and Interactive Digital Entertainment}},
  Vol.~\bibinfo{volume}{13}. \bibinfo{pages}{99--106}.
\newblock


\bibitem[Qadri et~al\mbox{.}(2023)]%
        {qadri2023ai}
\bibfield{author}{\bibinfo{person}{Rida Qadri}, \bibinfo{person}{Renee Shelby},
  \bibinfo{person}{Cynthia~L Bennett}, {and} \bibinfo{person}{Emily Denton}.}
  \bibinfo{year}{2023}\natexlab{}.
\newblock \showarticletitle{AI’s Regimes of Representation: A
  Community-centered Study of Text-to-Image Models in South Asia}. In
  \bibinfo{booktitle}{\emph{Proceedings of the 2023 ACM Conference on Fairness,
  Accountability, and Transparency}}. \bibinfo{pages}{506--517}.
\newblock


\bibitem[Queerinai et~al\mbox{.}(2023)]%
        {queerinai2023queer}
\bibfield{author}{\bibinfo{person}{Organizers~Of Queerinai},
  \bibinfo{person}{Anaelia Ovalle}, \bibinfo{person}{Arjun Subramonian},
  \bibinfo{person}{Ashwin Singh}, \bibinfo{person}{Claas Voelcker},
  \bibinfo{person}{Danica~J Sutherland}, \bibinfo{person}{Davide Locatelli},
  \bibinfo{person}{Eva Breznik}, \bibinfo{person}{Filip Klubicka},
  \bibinfo{person}{Hang Yuan}, {et~al\mbox{.}}}
  \bibinfo{year}{2023}\natexlab{}.
\newblock \showarticletitle{Queer In AI: A Case Study in Community-Led
  Participatory AI}. In \bibinfo{booktitle}{\emph{Proceedings of the 2023 ACM
  Conference on Fairness, Accountability, and Transparency}}.
  \bibinfo{pages}{1882--1895}.
\newblock


\bibitem[Raskin(1979)]%
        {raskin1979semantic}
\bibfield{author}{\bibinfo{person}{Victor Raskin}.}
  \bibinfo{year}{1979}\natexlab{}.
\newblock \showarticletitle{Semantic mechanisms of humor}. In
  \bibinfo{booktitle}{\emph{Annual Meeting of the Berkeley Linguistics
  Society}}, Vol.~\bibinfo{volume}{5}. \bibinfo{pages}{325--335}.
\newblock


\bibitem[Rauh et~al\mbox{.}(2022)]%
        {rauh2022characteristics}
\bibfield{author}{\bibinfo{person}{Maribeth Rauh}, \bibinfo{person}{John
  Mellor}, \bibinfo{person}{Jonathan Uesato}, \bibinfo{person}{Po-Sen Huang},
  \bibinfo{person}{Johannes Welbl}, \bibinfo{person}{Laura Weidinger},
  \bibinfo{person}{Sumanth Dathathri}, \bibinfo{person}{Amelia Glaese},
  \bibinfo{person}{Geoffrey Irving}, \bibinfo{person}{Iason Gabriel},
  {et~al\mbox{.}}} \bibinfo{year}{2022}\natexlab{}.
\newblock \showarticletitle{Characteristics of harmful text: Towards rigorous
  benchmarking of language models}.
\newblock \bibinfo{journal}{\emph{Advances in Neural Information Processing
  Systems}}  \bibinfo{volume}{35} (\bibinfo{year}{2022}),
  \bibinfo{pages}{24720--24739}.
\newblock


\bibitem[Ritchie(1999)]%
        {ritchie1999developing}
\bibfield{author}{\bibinfo{person}{Graeme Ritchie}.}
  \bibinfo{year}{1999}\natexlab{}.
\newblock \showarticletitle{Developing the Incongruity-Resolution Theory}.
\newblock \bibinfo{journal}{\emph{Institute for Communicating and Collaborative
  Systems}} (\bibinfo{year}{1999}).
\newblock


\bibitem[Rosa et~al\mbox{.}(2020)]%
        {rosa2020theaitre}
\bibfield{author}{\bibinfo{person}{Rudolf Rosa}, \bibinfo{person}{Ond{\v{r}}ej
  Du{\v{s}}ek}, \bibinfo{person}{Tom Kocmi}, \bibinfo{person}{David
  Mare{\v{c}}ek}, \bibinfo{person}{Tom{\'a}{\v{s}} Musil},
  \bibinfo{person}{Patr{\'\i}cia Schmidtov{\'a}}, \bibinfo{person}{Dominik
  Jurko}, \bibinfo{person}{Ond{\v{r}}ej Bojar}, \bibinfo{person}{Daniel Hrbek},
  \bibinfo{person}{David Ko{\v{s}}t'{\'a}k}, {et~al\mbox{.}}}
  \bibinfo{year}{2020}\natexlab{}.
\newblock \showarticletitle{THEaiTRE: Artificial intelligence to write a
  theatre play}.
\newblock \bibinfo{journal}{\emph{arXiv preprint arXiv:2006.14668}}
  (\bibinfo{year}{2020}).
\newblock


\bibitem[Rosa et~al\mbox{.}(2022)]%
        {rosa2022gpt}
\bibfield{author}{\bibinfo{person}{Rudolf Rosa}, \bibinfo{person}{Patr{\'\i}cia
  Schmidtov{\'a}}, \bibinfo{person}{Ond{\v{r}}ej Du{\v{s}}ek},
  \bibinfo{person}{Tom{\'a}{\v{s}} Musil}, \bibinfo{person}{David
  Mare{\v{c}}ek}, \bibinfo{person}{Saad Obaid}, \bibinfo{person}{Marie
  Nov{\'a}kov{\'a}}, \bibinfo{person}{Kl{\'a}ra Voseck{\'a}}, {and}
  \bibinfo{person}{Josef Dole{\v{z}}al}.} \bibinfo{year}{2022}\natexlab{}.
\newblock \showarticletitle{GPT-2-based Human-in-the-loop Theatre Play Script
  Generation}. In \bibinfo{booktitle}{\emph{Proceedings of the 4th Workshop of
  Narrative Understanding (WNU2022)}}. \bibinfo{pages}{29--37}.
\newblock


\bibitem[Santurkar et~al\mbox{.}(2023)]%
        {santurkar2023whose}
\bibfield{author}{\bibinfo{person}{Shibani Santurkar}, \bibinfo{person}{Esin
  Durmus}, \bibinfo{person}{Faisal Ladhak}, \bibinfo{person}{Cinoo Lee},
  \bibinfo{person}{Percy Liang}, {and} \bibinfo{person}{Tatsunori Hashimoto}.}
  \bibinfo{year}{2023}\natexlab{}.
\newblock \showarticletitle{Whose opinions do language models reflect?}. In
  \bibinfo{booktitle}{\emph{International Conference on Machine Learning}}.
  PMLR, \bibinfo{pages}{29971--30004}.
\newblock


\bibitem[Schmidtov{\'a} et~al\mbox{.}(2022)]%
        {schmidtova2022dialoguescript}
\bibfield{author}{\bibinfo{person}{Patr{\'\i}cia Schmidtov{\'a}},
  \bibinfo{person}{D{\'a}vid Javorsk{\`y}}, \bibinfo{person}{Christi{\'a}n
  Mikl{\'a}{\v{s}}}, \bibinfo{person}{Tom{\'a}{\v{s}} Musil},
  \bibinfo{person}{Rudolf Rosa}, {and} \bibinfo{person}{Ond{\v{r}}ej
  Du{\v{s}}ek}.} \bibinfo{year}{2022}\natexlab{}.
\newblock \showarticletitle{DialogueScript: Using Dialogue Agents to Produce a
  Script}.
\newblock \bibinfo{journal}{\emph{arXiv preprint arXiv:2206.08425}}
  (\bibinfo{year}{2022}).
\newblock


\bibitem[Shaotran et~al\mbox{.}(2023)]%
        {shaotran2023aligned}
\bibfield{author}{\bibinfo{person}{Ethan Shaotran}, \bibinfo{person}{Ido
  Pesok}, \bibinfo{person}{Sam Jones}, {and} \bibinfo{person}{Emi Liu}.}
  \bibinfo{year}{2023}\natexlab{}.
\newblock \showarticletitle{Aligned: A Platform-based Process for Alignment}.
\newblock \bibinfo{journal}{\emph{arXiv preprint arXiv:2311.08706}}
  (\bibinfo{year}{2023}).
\newblock


\bibitem[Sharples and y~P{\'e}rez(2022)]%
        {sharples2022story}
\bibfield{author}{\bibinfo{person}{Mike Sharples} {and}
  \bibinfo{person}{Rafael~P{\'e}rez y P{\'e}rez}.}
  \bibinfo{year}{2022}\natexlab{}.
\newblock \bibinfo{booktitle}{\emph{Story machines: How computers have become
  creative writers}}.
\newblock \bibinfo{publisher}{Routledge}.
\newblock


\bibitem[Sheng et~al\mbox{.}(2021)]%
        {sheng2021societal}
\bibfield{author}{\bibinfo{person}{Emily Sheng}, \bibinfo{person}{Kai-Wei
  Chang}, \bibinfo{person}{Prem Natarajan}, {and} \bibinfo{person}{Nanyun
  Peng}.} \bibinfo{year}{2021}\natexlab{}.
\newblock \showarticletitle{Societal Biases in Language Generation: Progress
  and Challenges}. In \bibinfo{booktitle}{\emph{Proceedings of the 59th Annual
  Meeting of the Association for Computational Linguistics and the 11th
  International Joint Conference on Natural Language Processing (Volume 1: Long
  Papers)}}. \bibinfo{pages}{4275--4293}.
\newblock


\bibitem[Simonton(2000)]%
        {simonton2000creativity}
\bibfield{author}{\bibinfo{person}{Dean~Keith Simonton}.}
  \bibinfo{year}{2000}\natexlab{}.
\newblock \showarticletitle{Creativity: Cognitive, personal, developmental, and
  social aspects.}
\newblock \bibinfo{journal}{\emph{American psychologist}} \bibinfo{volume}{55},
  \bibinfo{number}{1} (\bibinfo{year}{2000}), \bibinfo{pages}{151}.
\newblock


\bibitem[Srinivasan et~al\mbox{.}(2021)]%
        {srinivasan2021art}
\bibfield{author}{\bibinfo{person}{Ramya~Malur Srinivasan},
  \bibinfo{person}{Emily Denton}, \bibinfo{person}{Jordan~Jennifer Famularo},
  \bibinfo{person}{Negar Rostamzadeh}, \bibinfo{person}{Fernando Diaz}, {and}
  \bibinfo{person}{Beth Coleman}.} \bibinfo{year}{2021}\natexlab{}.
\newblock \showarticletitle{Art Sheets for Art Datasets}.
\newblock
\urldef\tempurl%
\url{https://openreview.net/pdf?id=K7ke_GZ_6N}
\showURL{%
\tempurl}


\bibitem[Stock and Strapparava(2005)]%
        {stock2005hahacronym}
\bibfield{author}{\bibinfo{person}{Oliviero Stock} {and} \bibinfo{person}{Carlo
  Strapparava}.} \bibinfo{year}{2005}\natexlab{}.
\newblock \showarticletitle{Hahacronym: A computational humor system}. In
  \bibinfo{booktitle}{\emph{Proceedings of the ACL Interactive Poster and
  Demonstration Sessions}}. \bibinfo{pages}{113--116}.
\newblock


\bibitem[Team et~al\mbox{.}({[n.\,d.]})]%
        {geminiteam2023gemini}
\bibfield{author}{\bibinfo{person}{Gemini Team}, \bibinfo{person}{Rohan Anil},
  \bibinfo{person}{Sebastian Borgeaud}, \bibinfo{person}{Yonghui Wu},
  \bibinfo{person}{Jean-Baptiste Alayrac}, \bibinfo{person}{Jiahui Yu},
  \bibinfo{person}{Radu Soricut}, \bibinfo{person}{Johan Schalkwyk},
  \bibinfo{person}{Andrew~M. Dai}, \bibinfo{person}{Anja Hauth},
  \bibinfo{person}{Katie Millican}, \bibinfo{person}{David Silver},
  \bibinfo{person}{Slav Petrov}, \bibinfo{person}{Melvin Johnson},
  \bibinfo{person}{Ioannis Antonoglou}, \bibinfo{person}{Julian Schrittwieser},
  \bibinfo{person}{Amelia Glaese}, \bibinfo{person}{Jilin Chen},
  \bibinfo{person}{Emily Pitler}, \bibinfo{person}{Timothy Lillicrap},
  {et~al\mbox{.}}} \bibinfo{year}{[n.\,d.]}\natexlab{}.
\newblock \bibinfo{title}{Gemini: A Family of Highly Capable Multimodal
  Models}.
\newblock
\newblock


\bibitem[Toplyn(2023)]%
        {toplyn2023witscript3}
\bibfield{author}{\bibinfo{person}{Joe Toplyn}.}
  \bibinfo{year}{2023}\natexlab{}.
\newblock \showarticletitle{Witscript 3: A Hybrid AI System for Improvising
  Jokes in a Conversation}.
\newblock \bibinfo{journal}{\emph{arXiv preprint arXiv:2301.02695}}
  (\bibinfo{year}{2023}).
\newblock


\bibitem[Touvron et~al\mbox{.}({[n.\,d.]})]%
        {touvron2023llama}
\bibfield{author}{\bibinfo{person}{Hugo Touvron}, \bibinfo{person}{Louis
  Martin}, \bibinfo{person}{Kevin Stone}, \bibinfo{person}{Peter Albert},
  \bibinfo{person}{Amjad Almahairi}, \bibinfo{person}{Yasmine Babaei},
  \bibinfo{person}{Nikolay Bashlykov}, \bibinfo{person}{Soumya Batra},
  \bibinfo{person}{Prajjwal Bhargava}, \bibinfo{person}{Shruti Bhosale},
  \bibinfo{person}{Dan Bikel}, \bibinfo{person}{Lukas Blecher},
  \bibinfo{person}{Cristian~Canton Ferrer}, \bibinfo{person}{Moya Chen},
  \bibinfo{person}{Guillem Cucurull}, \bibinfo{person}{David Esiobu},
  \bibinfo{person}{Jude Fernandes}, \bibinfo{person}{Jeremy Fu},
  \bibinfo{person}{Wenyin Fu}, {et~al\mbox{.}}}
  \bibinfo{year}{[n.\,d.]}\natexlab{}.
\newblock \bibinfo{title}{Llama 2: Open Foundation and Fine-Tuned Chat Models}.
\newblock
\newblock


\bibitem[Varshney(2023)]%
        {varshney2023decolonial}
\bibfield{author}{\bibinfo{person}{Kush~R Varshney}.}
  \bibinfo{year}{2023}\natexlab{}.
\newblock \showarticletitle{Decolonial AI Alignment: Vi$\backslash$'$\{$s$\}$
  esadharma, Argument, and Artistic Expression}.
\newblock \bibinfo{journal}{\emph{arXiv preprint arXiv:2309.05030}}
  (\bibinfo{year}{2023}).
\newblock


\bibitem[Veale(2021)]%
        {veale2021your}
\bibfield{author}{\bibinfo{person}{Tony Veale}.}
  \bibinfo{year}{2021}\natexlab{}.
\newblock \bibinfo{booktitle}{\emph{Your Wit is My Command: Building AIs with a
  Sense of Humor}}.
\newblock \bibinfo{publisher}{Mit Press}.
\newblock


\bibitem[Weidinger et~al\mbox{.}(2021)]%
        {weidinger2021ethical}
\bibfield{author}{\bibinfo{person}{Laura Weidinger}, \bibinfo{person}{John
  Mellor}, \bibinfo{person}{Maribeth Rauh}, \bibinfo{person}{Conor Griffin},
  \bibinfo{person}{Jonathan Uesato}, \bibinfo{person}{Po-Sen Huang},
  \bibinfo{person}{Myra Cheng}, \bibinfo{person}{Mia Glaese},
  \bibinfo{person}{Borja Balle}, \bibinfo{person}{Atoosa Kasirzadeh},
  {et~al\mbox{.}}} \bibinfo{year}{2021}\natexlab{}.
\newblock \showarticletitle{Ethical and social risks of harm from language
  models}.
\newblock \bibinfo{journal}{\emph{arXiv preprint arXiv:2112.04359}}
  (\bibinfo{year}{2021}).
\newblock


\bibitem[Weidinger et~al\mbox{.}(2023)]%
        {weidinger2023sociotechnical}
\bibfield{author}{\bibinfo{person}{Laura Weidinger}, \bibinfo{person}{Maribeth
  Rauh}, \bibinfo{person}{Nahema Marchal}, \bibinfo{person}{Arianna Manzini},
  \bibinfo{person}{Lisa~Anne Hendricks}, \bibinfo{person}{Juan Mateos-Garcia},
  \bibinfo{person}{Stevie Bergman}, \bibinfo{person}{Jackie Kay},
  \bibinfo{person}{Conor Griffin}, \bibinfo{person}{Ben Bariach},
  {et~al\mbox{.}}} \bibinfo{year}{2023}\natexlab{}.
\newblock \showarticletitle{Sociotechnical Safety Evaluation of Generative AI
  Systems}.
\newblock \bibinfo{journal}{\emph{arXiv preprint arXiv:2310.11986}}
  (\bibinfo{year}{2023}).
\newblock


\bibitem[West et~al\mbox{.}(2019)]%
        {west2019discriminating}
\bibfield{author}{\bibinfo{person}{Sarah~Myers West}, \bibinfo{person}{Meredith
  Whittaker}, {and} \bibinfo{person}{Kate Crawford}.}
  \bibinfo{year}{2019}\natexlab{}.
\newblock \showarticletitle{Discriminating systems}.
\newblock \bibinfo{journal}{\emph{AI Now}} (\bibinfo{year}{2019}),
  \bibinfo{pages}{1--33}.
\newblock


\bibitem[Winters(2021)]%
        {winters2021computers}
\bibfield{author}{\bibinfo{person}{Thomas Winters}.}
  \bibinfo{year}{2021}\natexlab{}.
\newblock \showarticletitle{Computers Learning Humor Is No Joke}.
\newblock \bibinfo{journal}{\emph{Harvard Data Science Review}}
  \bibinfo{volume}{3}, \bibinfo{number}{2} (\bibinfo{year}{2021}).
\newblock


\bibitem[Winters and Delobelle(2021)]%
        {winters2021survival}
\bibfield{author}{\bibinfo{person}{Thomas Winters} {and}
  \bibinfo{person}{Pieter Delobelle}.} \bibinfo{year}{2021}\natexlab{}.
\newblock \showarticletitle{Survival of the wittiest: Evolving satire with
  language models}. In \bibinfo{booktitle}{\emph{Proceedings of the Twelfth
  International Conference on Computational Creativity}}. Association for
  Computational Creativity (ACC), \bibinfo{pages}{82--86}.
\newblock


\bibitem[Winters and Mathewson(2019)]%
        {winters2019automatically}
\bibfield{author}{\bibinfo{person}{Thomas Winters} {and}
  \bibinfo{person}{Kory~W Mathewson}.} \bibinfo{year}{2019}\natexlab{}.
\newblock \showarticletitle{Automatically generating engaging presentation
  slide decks}. In \bibinfo{booktitle}{\emph{International Conference on
  Computational Intelligence in Music, Sound, Art and Design (Part of
  EvoStar)}}. Springer, \bibinfo{pages}{127--141}.
\newblock


\bibitem[Winters et~al\mbox{.}(2018)]%
        {winters2018automatic}
\bibfield{author}{\bibinfo{person}{Thomas Winters}, \bibinfo{person}{Vincent
  Nys}, {and} \bibinfo{person}{Daniel De~Schreye}.}
  \bibinfo{year}{2018}\natexlab{}.
\newblock \showarticletitle{Automatic joke generation: Learning humor from
  examples}. In \bibinfo{booktitle}{\emph{Distributed, Ambient and Pervasive
  Interactions: Technologies and Contexts: 6th International Conference, DAPI
  2018, Held as Part of HCI International 2018, Las Vegas, NV, USA, July
  15--20, 2018, Proceedings, Part II 6}}. Springer, \bibinfo{pages}{360--377}.
\newblock


\bibitem[Winters et~al\mbox{.}(2019)]%
        {winters2019towards}
\bibfield{author}{\bibinfo{person}{Thomas Winters}, \bibinfo{person}{Vincent
  Nys}, {and} \bibinfo{person}{Danny De~Schreye}.}
  \bibinfo{year}{2019}\natexlab{}.
\newblock \showarticletitle{Towards a general framework for humor generation
  from rated examples}. In \bibinfo{booktitle}{\emph{Proceedings of the 10th
  International Conference on Computational Creativity}}. Association for
  Computational Creativity, \bibinfo{pages}{274--281}.
\newblock
\urldef\tempurl%
\url{http://computationalcreativity.
  net/iccc2019/assets/iccc\_proceedings\_2019. pdf}
\showURL{%
\tempurl}


\bibitem[Workshop et~al\mbox{.}(2022)]%
        {workshop2022bloom}
\bibfield{author}{\bibinfo{person}{BigScience Workshop},
  \bibinfo{person}{Teven~Le Scao}, \bibinfo{person}{Angela Fan},
  \bibinfo{person}{Christopher Akiki}, \bibinfo{person}{Ellie Pavlick},
  \bibinfo{person}{Suzana Ili{\'c}}, \bibinfo{person}{Daniel Hesslow},
  \bibinfo{person}{Roman Castagn{\'e}}, \bibinfo{person}{Alexandra~Sasha
  Luccioni}, \bibinfo{person}{Fran{\c{c}}ois Yvon}, {et~al\mbox{.}}}
  \bibinfo{year}{2022}\natexlab{}.
\newblock \showarticletitle{Bloom: A 176b-parameter open-access multilingual
  language model}.
\newblock \bibinfo{journal}{\emph{arXiv preprint arXiv:2211.05100}}
  (\bibinfo{year}{2022}).
\newblock


\bibitem[Xu et~al\mbox{.}(2021)]%
        {xu2021detoxifying}
\bibfield{author}{\bibinfo{person}{Albert Xu}, \bibinfo{person}{Eshaan Pathak},
  \bibinfo{person}{Eric Wallace}, \bibinfo{person}{Suchin Gururangan},
  \bibinfo{person}{Maarten Sap}, {and} \bibinfo{person}{Dan Klein}.}
  \bibinfo{year}{2021}\natexlab{}.
\newblock \showarticletitle{Detoxifying Language Models Risks Marginalizing
  Minority Voices}. In \bibinfo{booktitle}{\emph{Proceedings of the 2021
  Conference of the North American Chapter of the Association for Computational
  Linguistics: Human Language Technologies}}. \bibinfo{pages}{2390--2397}.
\newblock


\bibitem[Yang et~al\mbox{.}(2022)]%
        {yang2022re3}
\bibfield{author}{\bibinfo{person}{Kevin Yang}, \bibinfo{person}{Yuandong
  Tian}, \bibinfo{person}{Nanyun Peng}, {and} \bibinfo{person}{Dan Klein}.}
  \bibinfo{year}{2022}\natexlab{}.
\newblock \showarticletitle{Re3: Generating Longer Stories With Recursive
  Reprompting and Revision}. In \bibinfo{booktitle}{\emph{Proceedings of the
  2022 Conference on Empirical Methods in Natural Language Processing}}.
  \bibinfo{pages}{4393--4479}.
\newblock


\bibitem[Yuan et~al\mbox{.}(2022)]%
        {yuan2022wordcraft}
\bibfield{author}{\bibinfo{person}{Ann Yuan}, \bibinfo{person}{Andy Coenen},
  \bibinfo{person}{Emily Reif}, {and} \bibinfo{person}{Daphne Ippolito}.}
  \bibinfo{year}{2022}\natexlab{}.
\newblock \showarticletitle{Wordcraft: Story Writing With Large Language
  Models}. In \bibinfo{booktitle}{\emph{27th International Conference on
  Intelligent User Interfaces}}. \bibinfo{pages}{841--852}.
\newblock


\bibitem[Zhou et~al\mbox{.}(2021)]%
        {zhou2021frequency}
\bibfield{author}{\bibinfo{person}{Kaitlyn Zhou}, \bibinfo{person}{Kawin
  Ethayarajh}, {and} \bibinfo{person}{Dan Jurafsky}.}
  \bibinfo{year}{2021}\natexlab{}.
\newblock \showarticletitle{Frequency-based distortions in contextualized word
  embeddings}.
\newblock \bibinfo{journal}{\emph{arXiv preprint arXiv:2104.08465}}
  (\bibinfo{year}{2021}).
\newblock


\bibitem[Ziegler et~al\mbox{.}(2019)]%
        {ziegler2019fine}
\bibfield{author}{\bibinfo{person}{Daniel~M Ziegler}, \bibinfo{person}{Nisan
  Stiennon}, \bibinfo{person}{Jeffrey Wu}, \bibinfo{person}{Tom~B Brown},
  \bibinfo{person}{Alec Radford}, \bibinfo{person}{Dario Amodei},
  \bibinfo{person}{Paul Christiano}, {and} \bibinfo{person}{Geoffrey Irving}.}
  \bibinfo{year}{2019}\natexlab{}.
\newblock \showarticletitle{Fine-tuning language models from human
  preferences}.
\newblock \bibinfo{journal}{\emph{arXiv preprint arXiv:1909.08593}}
  (\bibinfo{year}{2019}).
\newblock


\end{thebibliography}
	
	\appendix
	\section{Related Work on Computational Humour, AI and Comedy}
\label{sec:computational-comedy}

This section reviews computational humour literature. As foreshadowed in the Introduction (Sect. \ref{sec:fascination}) and discussed in the Section on the importance of context in humour and comedy (Sect. \ref{sec:human}), humour remains an elusive goal for AI.

\subsection{Humour generation, from template-based systems to prompting}

\citet{amin2020survey}, \citet{winters2021computers} and \citet{veale2021your} provide extensive surveys on the history of computational humour generation and early approaches, like hand-coded rules \citep{winters2018automatic} and templates for puns, riddles and acronyms \citep{binsted1994implemented,stock2005hahacronym}, or templates for automatically-generated slide decks for improvised ``Powerpoint Karaoke'' \citep{winters2019automatically}.


With the advent of LLMs, able to generate grammatically correct sentences in a given writing style, writers can focus on joke structure, guiding LLM-based systems to replicate such structures from examples of jokes \citep{winters2019towards}, or use evolutionary computing to evolve text to become more satirical \citep{winters2021survival}. More broadly, \citet{kaddour2023challenges} summarise recent work on using LLMs for creative applications such as (potentially humorous) story generation, including ``Recursive Reprompting and Revision'' \citep{yang2022re3} or interactive writing tools like ``WordCraft'' \citep{ippolito2022creative} and ``Dramatron'' \citep{mirowski2023co}.



\subsection{LLMs for comedy performance}


LLMs have been deployed since at least 2016 for live performance on stage, including in improvised comedy~\citep{mathewson2017improvised,mirowski2019human,branch2021collaborative}, short comical film scripts like \emph{Sunspring}, and song lyrics for musical comedy like \emph{Beyond the Fence}\footnote{\url{https://www.theguardian.com/stage/2016/feb/28/beyond-the-fence-review-computer-created-musical-arts-theatre-london}}. More recently, LLMs have been involved in (involuntarily) comical 
, absurdist productions of Prague-based company THEaiTRE~\citep{rosa2020theaitre,schmidtova2022dialoguescript,rosa2022gpt}, and semi-improvised comedy \emph{Plays By Bots}\footnote{Review: \url{https://12thnight.ca/2022/08/13/oh-no-bots-have-invaded-theatre-and-they-can-do-it-plays-by-bots-a-fringe-review/}}~\citep{mirowski2023co}. The success of these performances with AI primarily relied on the skills of the actors, who could invent subtext and add interpretation to AI-generated text \citep{mirowski2019human}.

\section{Participant questionnaire}
\label{app:questionnaire}

\subsection{{\tt Your past experience with AI systems in general.}}
\label{app:past}

{\tt
Please answer based on your past experience, before this workshop.
\begin{itemize}
    \item Have you used AI systems to write comedy material in the past? [Yes, and I used that material in some of my performances / Yes, but only to explore ideas / No]
    \item Have you used AI systems to generate content (including audio, video, text, etc.) for some of your performances? [Yes, and I operated the AI system myself / Yes, but someone else operated the AI system / No]
    \item In fewer than 10 words, why did you use AI in your artistic process? For example: trendy topic, glitch aesthetic, inspiration, etc.
    \item In fewer than 10 words, what is your background with AI? (Please do not give personally identifiable information)
    \item In few than 10 words, what is your performance background? For example: full-time comedian, dancer, years of experience, etc. (Please do not give personally identifiable information)
\end{itemize}
}

Note that the wording of these questions (for example: trendy topic, glitch aesthetic, inspiration) could be considered as potentially biasing, but we believed it was ultimately helpful to prompt study participants.

\subsection{{\tt Your experience with the AI system for writing comedy material}}
\label{app:experience}

{\tt
Please answer based on your interaction with the AI system in this morning's writing session.

\begin{itemize}
    \item I found the AI system helpful. [1 to 5]
    \item I felt like I was collaborating with the AI system. [1 to 5]
    \item I found it easy to write with the AI system. [1 to 5]
    \item I enjoyed writing with the AI system. [1 to 5]
    \item I was able to express my creative goals while writing with the AI system. [1 to 5]
    \item The comedy material written with the AI system feel unique. [1 to 5]
    \item I feel I have ownership over the comedy material written with the AI system. [1 to 5]
    \item I was surprised by the responses from the AI system. [1 to 5]
    \item I’m proud of the comedy material written with the AI system. [1 to 5]
\end{itemize}
}
In the questions above, 1 corresponds to {\tt Strongly disagree} and 5 to {\tt Strongly agree}.

\subsection{{\tt Creativity Support Index of the AI writing tool}}
\label{app:csi}

{\tt
This section contains questions that will allow us to compute a Creativity Support Index for the AI writing tool. Most of these questions overlap with those in the previous section.

\begin{itemize}
    \item The AI system allowed other people to work with me easily. [1 to 10]
    \item It was really easy to share ideas and designs with other people inside this system or tool. [1 to 10]
    \item I would be happy to use this system or tool on a regular basis. [1 to 10]
    \item I enjoyed using the system or tool. [1 to 10]
    \item It was easy for me to explore many different ideas, options, designs, or outcomes, using this system or tool. [1 to 10]
    \item The AI system was helpful in allowing me to track different ideas, outcomes, or possibilities. [1 to 10]
    \item I was able to be very creative while doing the activity inside this system or tool. [1 to 10]
    \item The system or tool allowed me to be very expressive. [1 to 10]
    \item My attention was fully tuned to the activity, and I forgot about the AI system that I was using. [1 to 10]
    \item I became so absorbed in the activity that I forgot about the AI system that I was using. [1 to 10]
    \item What I was able to produce was worth the effort I had to exert to produce it. [1 to 10]
    \item I was satisfied with what I got out of the system or tool. [1 to 10]
\end{itemize}
}
In the questions above, 1 corresponds to {\tt Strongly disagree} and 10 to {\tt Strongly agree}.

For the following question:
{\tt When writing comedy material, it is most important that I'm able to:}
the participant was shown two choices of response and asked to choose one of those two choices. There were six possible responses (listed below). Given that we consider pairs of 2 different responses at a time, there are $C_6^2 = \frac{6!}{4!2!} = 15$ unique pairwise choices.

{\tt
\begin{itemize}
    \item Be creative and expressive
    \item Become immersed in the activity
    \item Enjoy using the system or tool
    \item Explore many different ideas, outcomes, or possibilities
    \item Produce results that are worth the effort I put in
    \item Work with other people
\end{itemize}
}

All the questions listed in this section \ref{app:csi} are directly taken from \citep{cherry2014quantifying} and the NASA Task Load Index \citep{hart1988development}, and are used as is for easy of reproduction of results. Note that the questions pertaining to human-human collaboration are less pertinent to our study.

\subsection{{\tt Free-form questions about the AI system for writing.}}
\label{app:free}

{\tt
\begin{itemize}
    \item What is one thing that the AI system did well?
    \item What is one improvement for the AI system?
    \item Please provide any comments, reflections, or open questions that came up for you during the writing session.
    \item Please provide any other comments, reflections, or open questions that came up for you when answering this survey.
\end{itemize}
}

\section{Focus Group Questions}

\subsection{{\tt Qualitative questions about the specific writing task}}

{\tt
\begin{itemize}
    \item Did you find any of the generated outputs helpful? If so, could you recall one output that was usable and explain in what way it helped you write?
    \item Could you recall one generated output that was not usable, and explain why?
    \item How do you think the generated output differed from what you would find in some of the resources you use often, e.g., Wikipedia, Google search, other artists' material?
    \item How did the prompts we suggested differ from your custom prompts?
    \item Comment on the types of comedy that you managed to generated with the AI tools.
    \item Were any of the generated outputs that were presented offensive/inappropriate in some way? If so, what did you think of these?
    \item What made you decide to stop generating outputs?
    \item Did you have any concerns about ownership or agency when generating outputs?
\end{itemize}
}

\subsection{{\tt General discussion points about AI tools for comedy}}

{\tt
\begin{itemize}
    \item What does your comedy writing process look like? How is this process of working with AI different from working alone or working with other comedians?
    \item What can you say about the stereotypes in the outputs generated by the AI writing tool? How do they relate to or differ from stereotypes in human written comedy?
    \item Is humor and comedy evolving over time? Should AI tools adapt to societal changes in a similar way?
    \item Does the computer have a "voice"? How would you compare it with the voice of a human comedian? What are the consequences for writing comedy about identity?
    \item What is your stance on moderation? Should the output of AI tools be allowed to be edgy? Can and should the output of AI tools be moderated?
    \item How does context shape the meaning of comedy? Who has the responsibility for the comedy material?
    \item When is it acceptable to use other comedians', or in general, other artists' work? What about the outputs of an AI writing tool?
    \item What importance do you attach to the text vs. to the delivery of that text? Could a human comedian make AI-generated content better?
    \item Can comedy and humour be quantified and measured?
\end{itemize}
}

Note that the leading questions in the focus group could be considered as biasing the respondents. We chose this formulation to be able to address the topics of interest listed in Section \ref{sec:expected}.

\subsection{{\tt Debriefing questions}}

{\tt
\begin{itemize}
    \item Is there anything you’d like to share that I didn’t ask about?
    \item Is there anything you’d like to know or ask me?
\end{itemize}
}
	
\end{document}